\pdfoutput=1

\documentclass[11pt]{article}
\usepackage{hyperref}

\usepackage[]{EMNLP2023}

\usepackage{times}
\usepackage{latexsym}

\usepackage[T1]{fontenc}

\usepackage[utf8]{inputenc}

\usepackage{microtype}

\usepackage{inconsolata}

\usepackage{graphicx}
\usepackage{multirow}
\usepackage{multicol}
\usepackage{tabularx}
\usepackage{varwidth}
\usepackage{adjustbox,lipsum}
\usepackage{booktabs}
\usepackage{color,soul}
\usepackage{xcolor}
\PassOptionsToPackage{dvipsnames}{xcolor}
\usepackage{pdfpages}
\usepackage{subcaption}
\usepackage{bm}

\newcommand{\secref}[1]{\S\ref{#1}}
\newcommand{\dataset}{{\sc CovidET-Appraisals}}
\newcommand{\CovidET}{{\sc CovidET}}
\newcommand{\bftab}{\fontseries{b}\selectfont}

\usepackage[textsize=scriptsize]{todonotes}

\title{Evaluating Subjective Cognitive Appraisals of Emotions \\ from Large Language Models}

\author{
    \textbf{Hongli Zhan}$^1$\quad\textbf{Desmond C. Ong}$^2$\quad\textbf{Junyi Jessy Li}$^1$\\
    $^1$Department of Linguistics, The University of Texas at Austin\\$^2$Department of Psychology, The University of Texas at Austin\\
    \texttt{\{honglizhan, desmond.ong, jessy\}@utexas.edu}\\
}

\begin{document}
\maketitle

\begin{abstract}
The emotions we experience involve complex processes; besides physiological aspects, research in psychology has studied \emph{cognitive appraisals} where people assess their situations subjectively, according to their own values \cite{Scherer2005WhatAE}. Thus, the same situation can often result in different emotional experiences. While the \emph{detection} of emotion is a well-established task, there is very limited work so far on the automatic prediction of cognitive appraisals. This work fills the gap by presenting \dataset{}, the most comprehensive dataset to-date that assesses 24 appraisal dimensions, each with a natural language rationale, across 241 Reddit posts. \dataset{} presents an ideal testbed to evaluate the ability of large language models --- excelling at a wide range of NLP tasks --- to automatically assess and explain cognitive appraisals. We found that while the best models are performant, open-sourced LLMs fall short at this task, presenting a new challenge in the future development of emotionally intelligent models. \emph{We release our dataset at \url{https://github.com/honglizhan/CovidET-Appraisals-Public}.}
\end{abstract}

\section{Introduction}

\begin{figure}[t]
    \centering
    \includegraphics[width=\columnwidth]{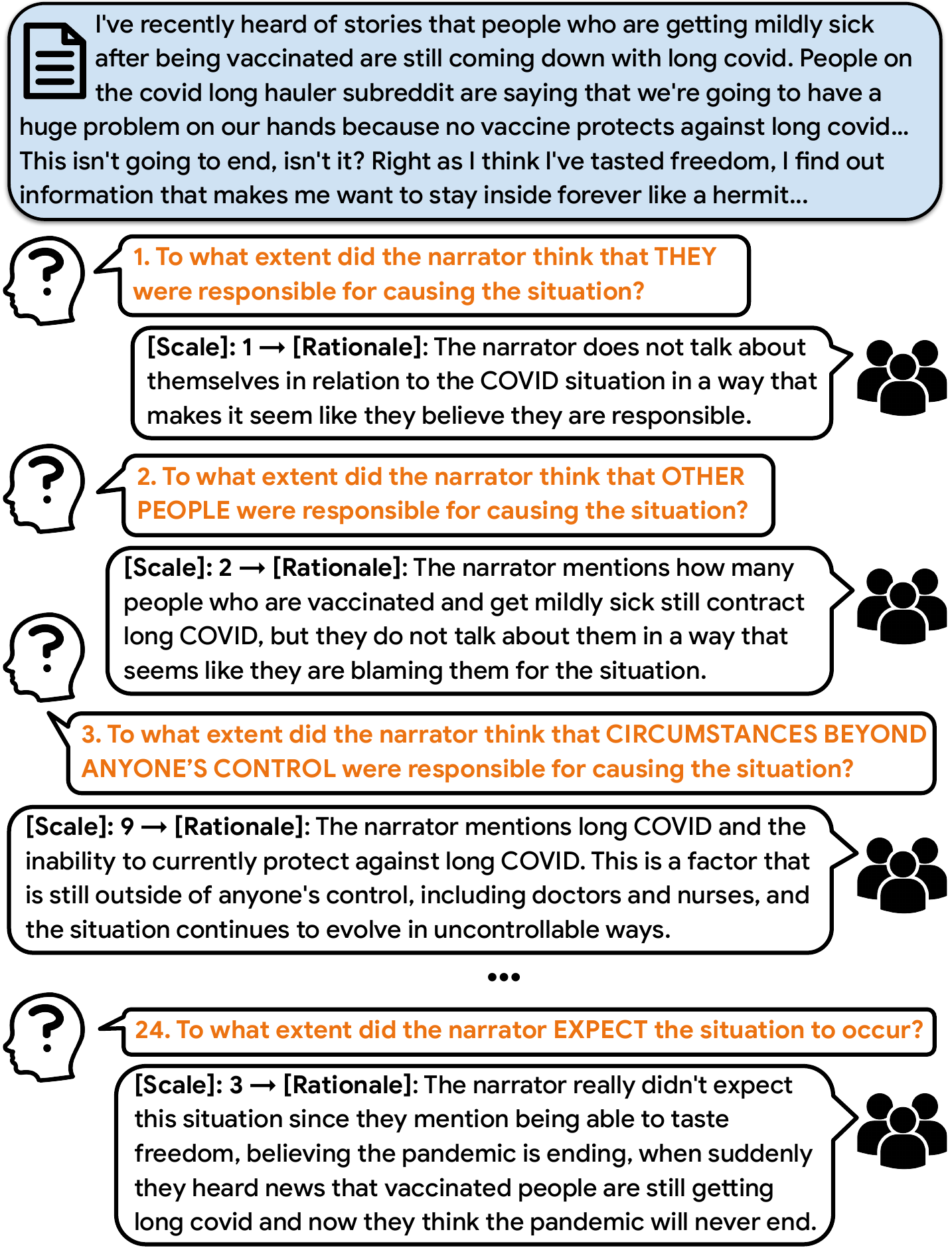}
    \caption{An example from \dataset{}. The fact that the narrator is blaming nobody but circumstances beyond anyone's control for causing long-COVID contributes to their feeling of \textit{sadness}. We showcase an annotation together with LLMs' responses in Appendix \secref{appendix:dataset_example}.}
    \label{fig:dataset-example}
\end{figure}

Emotions constitute a crucial aspect of people's lives, and understanding them has a profound impact on improving public mental health problems as well as policy-making \cite{Choudhury2014MentalHD, gjurkovic-snajder-2018-reddit, arora_role_2021, UBAN2021480}. The emotions we experience involve complex processes: the same situation can often result in different emotional experiences, based on an individual's subjective evaluations. These are called \textit{cognitive appraisals}, and have been extensively studied in psychology through theoretical, behavioral, and hand-coded studies \cite{arnold1960emotion, lazarus1966psychological, lazarus1980emotions, roseman1984cognitive, scherer1984nature, smith1985patterns, weiner1985attributional, clore2000cognition, roseman2001appraisal, scherer2001appraisal, ellsworth2003appraisal, sander2005systems, ong2015cognition, Ong2019ModelingEI, ortony2022cognitive, YeoOng2023Appraisals}. For instance, being fired from a job, if judged to be due to one's own controllable mistakes, could result in regret; if evaluated to be unfair and due to someone else's intentional actions, would make one feel angry; and if appraised to be leaving a toxic work environment, could instead result in relief and even happiness. \textbf{The different dimensions along which people subjectively interpret or \emph{appraise} the situation characterizes the specific emotions they feel}~\cite{moors-appraisal-2013}.

Although emotion \emph{detection} is a well-established NLP task \cite{strapparava-mihalcea-2007-semeval, mihalcea-strapparava-2012-lyrics, Wang2012HarnessingT, lei-online-news-2014, abdul-mageed-ungar-2017-emonet, khanpour-caragea-2018-fine, liu-etal-2019-dens, sosea-caragea-2020-canceremo, demszky-etal-2020-goemotions, desai-etal-2020-detecting, sosea-etal-2022-emotion}, it mostly involves classification from text to emotion labels directly, skipping the appraisal step that is necessary to interpret why the emotion is experienced by an individual in a particular event. Hence, we do not yet have a data-driven understanding of these cognitive appraisals in textual data. Yet recent work has started to show its necessity: \citet{hofmann-etal-2020-appraisal} showed that appraisals are informative for an emotion detection model; \citet{zhan-etal-2022-feel} further recognized appraisals to be an integral part of emotion triggers, though appraisals were not explicit in their work.

This work aims at construing an empirical, explicit understanding of \emph{perceived} cognitive appraisals in human readers and large language models (LLMs) alike, via a comprehensive $24$ dimensions, along with their corresponding natural language rationales. 
A language model's capability of assessing cognitive appraisals reflects a more nuanced understanding of emotions, where it could contextualize individual subjectivity in responses to the same situation, while offering explanations  (``they are feeling [\textit{emotion}] because of [\textit{appraisal}]''). This could be groundwork for emotional support agents, e.g., one capable of positive reframing \cite{ziems-etal-2022-inducing} or producing empathetic responses.

We first introduce \dataset{}, a dataset of $24$ appraisal dimensions annotated across $241$ Reddit posts sourced from \citet{zhan-etal-2022-feel} about COVID-19. Each post was manually annotated with $24$ appraisal dimensions from a recent meta-analysis covering all appraisal dimensions proposed and studied in the literature \cite{YeoOng2023Appraisals}. For each appraisal dimension, annotators not only rated the extent to which they perceived the narrator is experiencing the said dimension, but also provided a \emph{rationale} in their own language to justify their rating selection. An example from \dataset{} is shown in Figure \ref{fig:dataset-example}.

\dataset{} serves as an ideal testbed to evaluate the capability of a model to uncover implicit information for emotion understanding. Benchmarking on \dataset{}, we evaluate the performance of LLMs to (1) provide Likert-scale ratings for the appraisal dimensions; and (2) generate natural language rationales for their ratings. The elicitation of the rationales can be seen as a way of probing \cite{le-scao-rush-2021-many,gu-etal-2022-dream}, where we prefix a question with an elaborated situation. We evaluate a range of LLMs, including ChatGPT, Flan-T5~\cite{flan-t5-xxl}, Alpaca~\cite{alpaca}, Dolly~\cite{DatabricksBlog2023DollyV2}. With an extensive human evaluation of the natural language rationales from LLMs as well as our annotators, we find that ChatGPT performs on par with (and in some cases better than) human-annotated data; this opens a new avenue of investigation to improve its performance on emotion-related tasks~\cite{kocon2023chatgpt}. In comparison, other open-sourced LLMs fall short on this task, presenting a new challenge in the future development of emotionally intelligent open models.

\emph{We publicly release our annotated dataset \dataset{}, model outputs, and our human evaluation data at \url{https://github.com/honglizhan/CovidET-Appraisals-Public}.}

\section{Background and Related Work}

\paragraph{Cognitive Appraisal Theories.}
The cognitive appraisal theories of emotion state that emotions arise from an individual's subjective understanding and interpretation of situations that hold personal importance for their overall well-being \cite{arnold1960emotion, lazarus1966psychological, lazarus1980emotions, roseman1984cognitive, scherer1984nature, smith1985patterns, weiner1985attributional, clore2000cognition, roseman2001appraisal, scherer2001appraisal, sander2005systems, ortony2022cognitive}. In practical terms, people interpret and appraise situations along a range of different dimensions, and it is the specific manner in which they appraise their situations that give rise to the distinct emotions they experience. The primary focus of cognitive appraisal theories of emotions revolves around the identification of these appraisal dimensions that are associated with specific emotional experiences and how these dimensions contribute to distinguishing between different emotional states \cite{lazarus1993psychological, roseman1996appraisal, scherer2001appraisal, moors2010theories, scherer2019emotion}.

While appraisal theorists agree on the importance of motivationally-relevant appraisals in triggering emotions, they have not reached a consensus on the specific appraisal dimensions that play a significant role in this process \cite{YeoOng2023Appraisals}. Various theories have put forth distinct sets of appraisal dimensions that are considered crucial in triggering and distinguishing emotions. From prior literature, \citet{YeoOng2023Appraisals} identified and assembled a taxonomy of all appraisal dimensions that have been studied, and produced a condensed list of $24$ cognitive appraisal dimensions which we focus on in this paper.

\paragraph{Cognitive Appraisals in NLP.}
Appraisals provide the necessary computational structure allowing for the distillation of real-life situations that depend on a multitude of factors into a (large but) finite set of appraisal dimensions \cite{ong2015cognition}. Despite its importance, however, few works have explored the implications of cognitive appraisals on emotions in NLP. \citet{hofmann-etal-2020-appraisal} experimented with a small set of cognitive appraisal dimensions (including \textit{attention}, \textit{certainty}, \textit{effort}, \textit{pleasantness}, \textit{responsibility}, \textit{control}, and \textit{circumstance}) to assist the automatic detection of emotions in text, and found that accurate predictions of appraisal dimensions boost emotion classification performance. They introduced a dataset of $1,001$ sentences following the template ``I feel [\textit{emotion}], when ...'' (average sentence length: 27 tokens).
In comparison, our work covers a much wider range of $24$ appraisal dimensions found in prior literature, over lengthy (176 tokens on average) Reddit posts that were natural and emotionally charged.
We also collect natural language rationales as a key contribution to reveal human's in-depth understanding of such cognitive appraisals in context.

Recent studies \cite{zhan-etal-2022-feel, sosea-etal-2023-unsupervised} acknowledged both \emph{what happened and how one appraised the situation} as inherent components of emotion triggers, although the appraisal of events was not explicit in their work. Instead we provide datasets and perform evaluation on appraisals explicitly, such that language models can build on this work to achieve a comprehensive and explicit understanding of cognitive appraisals from written text.

\paragraph{LLMs on Emotion-Related Tasks.}
Autoregressive LLMs have been explored extensively in emotion-related tasks such as sentiment analysis \cite{zhong2023chatgpt, qin2023chatgpt, susnjak2023applying}, emotion recognition \cite{Koco__2023}, disclosing the representation of human emotions encapsulated in LLMs \cite{li2023human}, and interpreting mental health analysis \cite{yang2023interpretable}. However, few have tapped into the understanding of cognitive appraisals of emotions innate in LLMs. In this work, we dive into the extent to which LLMs comprehend the profound cognitive appraisals underlying emotions in situations, and further elicit natural language rationales from the language models to disclose the reason behind such predictions from the otherwise baffling black-box LLMs \cite{Gilpin2018ExplainingEA}. Aligning with \citet{marasovic-etal-2020-natural} who performed human evaluation on rationales generated by GPT, we additionally perform an in-depth human evaluation of the rationales from human annotators and LLMs alike on the novel task of providing natural language explanations for cognitive appraisals of situations that underlie narrators' emotional experiences.

\section{The \dataset{} Dataset}\label{sec:dataset}

\dataset{} contains $241$ Reddit posts sampled from the \CovidET{} dataset \cite{zhan-etal-2022-feel}, where the Reddit posts are sourced from \texttt{r/COVID19\_support}. Each post is manually annotated with one or more of the $7$ emotions: \textit{anger}, \textit{anticipation}, \textit{joy}, \textit{trust}, \textit{fear}, \textit{sadness}, and \textit{disgust}. The $241$ posts in \dataset{} have an average of $175.82$ tokens and $2.67$ emotions per post. From \citet{YeoOng2023Appraisals}'s work, we identify $24$ cognitive emotion appraisal dimensions (Table \ref{tab:dimension-labels}). We provide the instructions given to the annotators (including the full questions for each of these 24 dimensions) in Appendix \secref{appendix:annotation-framework}.

\begin{table}[t]
  \centering
  \small
  \adjustbox{max width=0.8\columnwidth}{
    \begin{tabular}{rll}
    \toprule
        ID & Abbrv. & Reader-Friendly Labels \\
    \midrule
    1     & \textit{srsp} & \textit{Self-responsibility} \\
    2     & \textit{orsp} & \textit{Other-responsibility} \\
    3     & \textit{crsp} & \textit{Circumstances-responsibility} \\
    4     & \textit{pfc} & \textit{Problem-focused coping} \\
    5     & \textit{grlv} & \textit{Goal Relevance} \\
    6     & \textit{attn} & \textit{Attentional activity} \\
    7     & \textit{efc} & \textit{Emotion-focused coping} \\
    8     & \textit{scrl} & \textit{Self-Controllable} \\
    9     & \textit{ocrl} & \textit{Other-Controllable} \\
    10    & \textit{ccrl} & \textit{Circumstances-Controllable} \\
    11    & \textit{prd} & \textit{Predictability} \\
    12    & \textit{thr} & \textit{Threat} \\
    13    & \textit{pls} & \textit{Pleasantness} \\
    14    & \textit{crt} & \textit{Certainty} \\
    15    & \textit{gcnd} & \textit{Goal Conduciveness} \\
    16    & \textit{fair} & \textit{Fairness} \\
    17    & \textit{fex} & \textit{Future expectancy} \\
    18    & \textit{csn} & \textit{Consistency with social norms} \\
    19    & \textit{loss} & \textit{Loss} \\
    20    & \textit{fml} & \textit{Familiarity} \\
    21    & \textit{eff} & \textit{Effort} \\
    22    & \textit{chl} & \textit{Challenge} \\
    23    & \textit{civ} & \textit{Consistency with internal values} \\
    24    & \textit{exp} & \textit{Expectedness} \\
    \bottomrule
    \end{tabular}}
  \caption{The 24 appraisal dimensions and their abbreviations we used throughout this paper. See Appendix \secref{appendix:annotation-framework} for full questions for each dimension, and Figure \ref{fig:dataset-example} for an example of how the items for 1: \textit{self-responsibility}, 2: \textit{other-responsibility}, 3: \textit{circumstances-responsibility}, and 24: \textit{expectedness} were framed.}
  \label{tab:dimension-labels}
\end{table}

\paragraph{Annotators.}
We recruited $2$ linguistics students at a university to work on our annotation task; both of them are native speakers of English. Both annotators underwent training using a set of posts already annotated by our group. Throughout the annotation, we monitored the inter-annotator agreement and provided feedback on their work.

\paragraph{Instructions.}
Given a Reddit post from \CovidET{}, annotators are asked to judge $24$ emotion appraisal dimensions pertaining to how the narrator feels about and views the situation that they are going through (e.g., whether the narrator feels the situation they are in is something they could control). For each appraisal dimension, annotators need to select a Likert rating on the scales of $1$ to $9$. A ``\textit{not mentioned}'' (NA) option is provided in case the dimension being asked is absent in the given post. In addition, we also ask the annotators to provide rationales for their ratings in the form of \emph{natural language explanations}.

On average, our trained annotators spent around $30$ minutes to complete the annotation of one post. Owing to the immense effort involved, we doubly annotate 40 posts to measure inter-annotator agreement while leaving the rest annotated by one annotator.

\begin{figure}[t]
    \centering
    \includegraphics[width=\columnwidth]{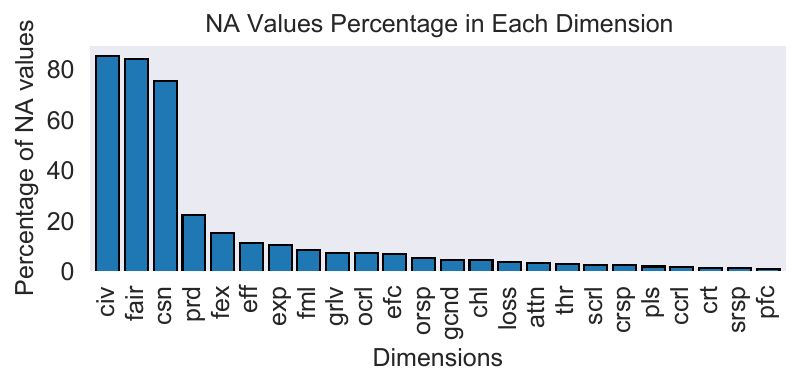}
    \caption{Percentage of \textit{``not mentioned''} labels in each dimension in \dataset{}.}
    \label{fig:percentage_na}
\end{figure}

\paragraph{Post-Processing and Aggregation.} 
Given a fixed topic (COVID-19 in our case), it is highly likely that certain dimensions frequently don't apply~\cite{YeoOng2023Appraisals}. This can be seen in Figure \ref{fig:percentage_na} which plots the percentage of NA labels: dimensions such as \textit{civ} (consistency with internal values), \textit{fair} (fairness), and \textit{csn} (consistency with social norms) contain mostly NA labels (around $80\%$). Therefore, we remove these dimensions from subsequent analyses and evaluations of the dataset. \textbf{This results in a total of 21 \emph{applicable} appraisal dimensions in \dataset{}.}

We collected $241$ posts in total. For the subset of $40$ posts that are doubly annotated, we aggregate the Likert-scale ratings by taking the mean of each post's ratings for each appraisal dimension (if an annotator labels a dimension as NA, we then exclude the particular dimension of that post that they annotate). In terms of the rationales, we consider both rationales as ground truth references and use multi-reference metrics in our experiments.

\paragraph{Inter-Annotator Agreement.}\label{paragraph:dataset-scale-agreement} 
We report inter-annotator agreement on the Likert-scale ratings. Since there is no reliable, automatic way to evaluate natural language rationales (as discussed in \secref{sec:analysis}), we evaluate them with human validation in \secref{subsec:human_eval}.

To measure the agreement for selecting the NA label, we average the Fleiss' Kappa values \cite{fleiss-1971, randolph2005free} across \emph{all} $24$ appraisal dimensions, yielding a value of $0.769$ indicating substantial agreement~\cite{artstein-poesio-2008-survey}.

For the $1$-$9$ Likert-scale ratings, we report on the $21$ applicable dimensions: (1) Spearman's $\rho$ between our two annotators, calculated per dimension then averaged across all dimensions; (2) Krippendorff's alpha (using interval distance) \cite{krippendorff_content_1980}; and (3) mean absolute difference (\emph{abs.\ delta}). Here the agreement is calculated if neither annotator gave a NA judgment. Krippendorff's alpha yields a value of $0.647$ indicating substantial agreement~\cite{artstein-poesio-2008-survey}. The average Spearman's correlation is $0.497$ with significance, and the absolute delta values also have a small mean of $1.734$. These measures indicate that while the task is subjective, annotators do align with each other with only a small difference compared to the scale of ratings ($1$-$9$). Agreement values differ by dimension, which we showcase in Appendix~\ref{appendix:inter-annotator-per-dimension}.

\section{Dataset Analysis}\label{sec:analysis}
\begin{figure}
    \centering
    \includegraphics[width=\columnwidth]{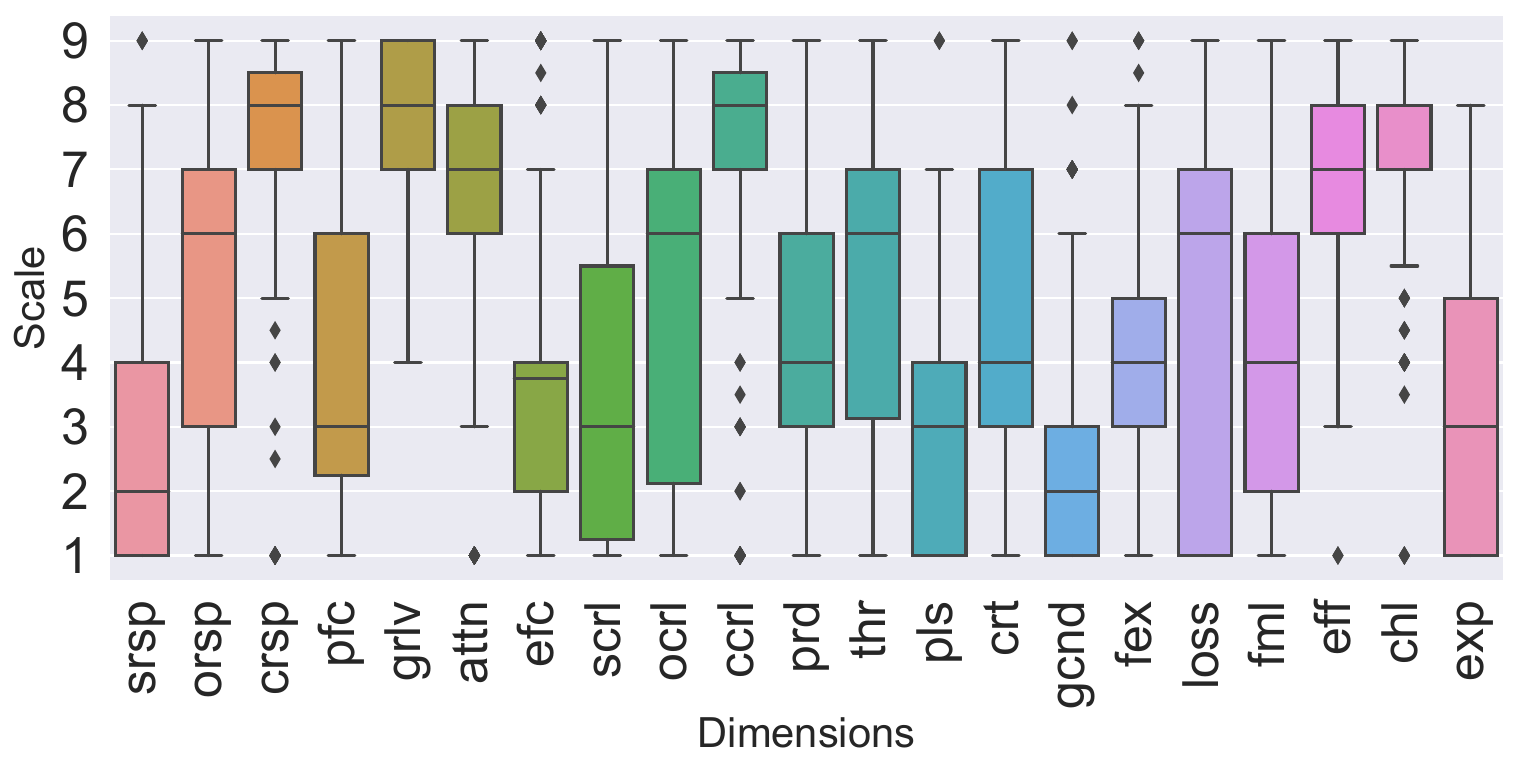}
    \caption{Distribution of the ratings for each dimension.}
    \label{fig:dimension_distribution_boxplot}
\end{figure}

\paragraph{How do the scales distribute across dimensions and emotions?}
The distribution of the Likert-scale ratings is shown in Figure \ref{fig:dimension_distribution_boxplot}. The ratings for some dimensions are consistent (e.g., dimensions \textit{crsp} (circumstances-responsibility), \textit{ccrl} (circumstances-controllable), and \textit{chl} (challenge)), whereas for some other dimensions, the ratings have higher variance (e.g., dimensions \textit{ocrl} (other-controllable) and \textit{loss}).

\begin{figure}[t]
    \centering
    \includegraphics[width=\columnwidth]{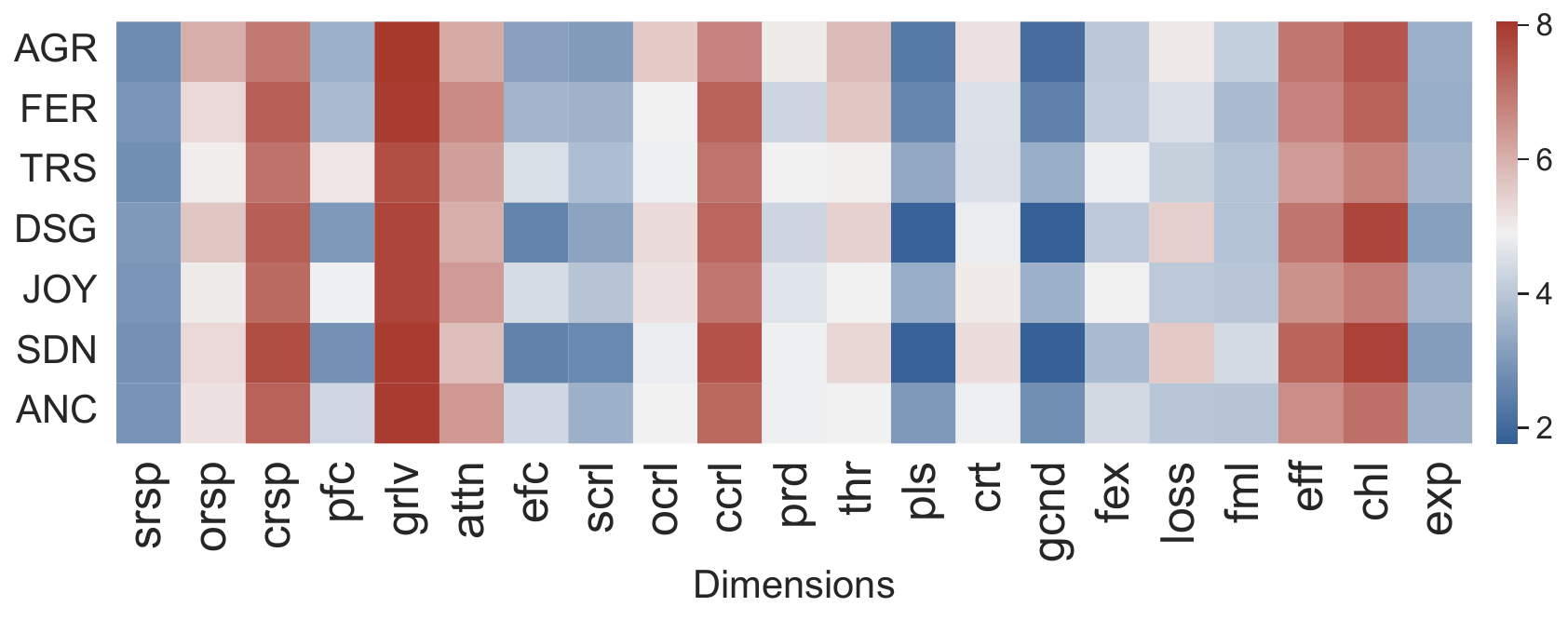}
    \caption{Mean Likert-scale ratings for each dimension in each emotion.}
    \label{fig:emotion_dimension_distribution}
\end{figure}

We analyze the connections between our Likert-scale annotations and \CovidET{}'s emotion annotations. Figure \ref{fig:emotion_dimension_distribution} shows the mean Likert-scale rating for each dimension within each post with respect to the perceived emotion. While it is evident that most dimensions show consistency (the posts are all related to COVID-19), some emotions stand out distinctly in particular dimensions. For example, \textit{trust} and \textit{joy} have higher Likert-scale ratings on dimensions \textit{pfc} (problem-focused coping) and \textit{gcnd} (goal conduciveness) compared to other emotions, suggesting the inter-correlation between these appraisal dimensions with positive emotions. We further explore whether appraisal dimensions alone are indicative of perceived emotions already annotated in {\sc CovidET} in Appendix \secref{appendix:dimension-feature-logistic-regression}.

\paragraph{What are the characteristics of the natural language rationales?}\label{paragraph:human_rationale_extractiveness} 
On average, each rationale is $1.2$ sentences (std.dev $= 0.4$) and $28.9$ tokens (std.dev $= 10.0$) long. Following \citet{marfurt-henderson-2021-sentence}, we also measure the abstractiveness of the rationales from our human annotators by calculating the percentage of novel bigrams in the rationales with respect to the Reddit posts and instructions (i.e., evaluating a specific appraisal dimension) that the annotators were given. As shown in Table \ref{tab:llm_results-rationale}, our human annotators attain a \% of novel bigrams of $86.7\%$, indicating a high abstractiveness. We showcase the most prominent topics extracted from the annotated rationales using Latent Dirichlet Allocation (LDA) \cite{10.5555/944919.944937} in Appendix \secref{appendix:lda}.

\paragraph{Are rationales repetitive?}

\begin{table}[t]
\small
  \centering
  \adjustbox{max width=\columnwidth}{
    \begin{tabular}{l||c|c|c}
    \toprule
    {} & \multicolumn{3}{c}{\textsc{Rationale}}\\
    {} & \multicolumn{1}{c|}{\textsc{BLEU-4}} & \multicolumn{1}{c|}{\textsc{ROUGE-L}} & \textsc{BERTSc}\\
    \midrule
    {\textsc{Annotators}} & $0.042$ & $0.253$ & \bftab 0.357 \\
    {\textsc{Baseline-P}} & \bftab 0.060 & \bftab 0.261 & $0.336$ \\
    {\textsc{Baseline-D}} & $0.059$ & $0.247$ & $0.332$ \\
    \bottomrule
    \end{tabular}}
  \caption{Automatic measures of similarity on the natural language rationales of \dataset{}. \textsc{Baseline-P} denotes ``baseline (same dimension, \textit{different posts})'', and \textsc{Baseline-D} denotes ``baseline (same post, \textit{different dimensions})''.}
  \label{tab:inter-agreement-rationale}
\end{table}

We also look into automatic measures of similarity to assess how much rationales from different annotators, or from different dimensions/posts, differ from one another. Specifically, we calculate BLEU-4~\cite{papineni-etal-2002-bleu}, ROUGE-L \cite{lin-2004-rouge}, and re-scaled BERTScore \cite{DBLP:journals/corr/abs-1904-09675} between our two annotators' rationales. We establish $2$ random baselines for comparison: (1) rationales of the same dimension from different posts; (2) rationales from different dimensions within the same post. In each case we report similarity between $3$ randomly sampled rationales and the annotated ones.

Table \ref{tab:inter-agreement-rationale} shows that the textual similarity in all conditions are somewhat low; the BLEU and ROUGE scores show that there is very little lexical overlap, although BERTScore shows higher semantic similarity between two annotators for the same dimension within the same post. Upon closer inspection, we observe that these commonly used automatic measures do not adequately capture semantic similarity in our dataset (see Appendix~\secref{appendix: semantic-similarity-example} for an example). This adds to the challenge of evaluating rationales; as a result, we resort to the human evaluation in \secref{subsec:human_eval}.

\section{Can LLMs understand emotional appraisals?}
\dataset{} provides an ideal testbed that evaluates models' performance on predicting both the Likert ratings, as well as their natural language explanations. Using \dataset{}, we evaluate the zero-shot performance of LLMs in an attempt to evaluate their innate ability to comprehend emotional appraisals from social media text without in-context learning.

\paragraph{Models.}
We evaluate the following instruction-tuned LLMs\footnote{While we have also experimented with non-instruction-tuned LLMs (including GPT-3 davinci and LLaMA (7B and 13B), they largely fail to generate sensible outputs for this task. We showcase examples of responses from non-instruction-tuned models in Appendix \secref{appendix:dataset_example}. For these reasons, we do not include their results in this paper.}: \textbf{1) ChatGPT}, i.e., GPT-3.5-Turbo; \textbf{2) FLAN-T5-XXL (11B)} \cite{flan-t5-xxl}, which is the instruction fine-tuned version of T5 \cite{2020t5}; \textbf{3) Alpaca (7B, 13B)} \cite{alpaca} is fine-tuned from LLaMA (7B and 13B) \cite{touvron2023llama} on $52$K instruction-following examples created with GPT text-davinci-003 in the manner of self-instruct \cite{wang2023selfinstruct}; \textbf{4) Dolly-V2 (7B, 12B)} \cite{DatabricksBlog2023DollyV2} is an instruction-tuned LLM trained on \textasciitilde$15$k demonstrations consisting of both instructions and responses.

\paragraph{Prompts and Setup.} 
The templates for prompting the LLMs are shown in Appendix Figure \ref{fig:prompt_templates}. After extensive experimentation, we found that only ChatGPT is able to generate both a rating and a rationale with a single prompt; this type of ``1-step'' prompting leads to ill-formed responses for other models. Thus, for models other than ChatGPT, we instead use a pipeline or ``2-step'' prompting similar to the strategy used in \citet{press2022measuring}: we first elicit the rating for the appraisal dimension, then conditioned on the response for the rating we further elicit the rationale for the selection.

We carry out all our experiments on $4$ Nvidia A$40$ GPUs. We use the HuggingFace Transformers \cite{wolf-etal-2020-transformers} library for model inference. We set the temperature value of all models to $0.1$.\footnote{We experimented with higher temperatures on a validation set consisting of $10$ Reddit posts annotated by our group which are not included in \dataset{}, and the models yielded worse and more unstable performance.} To enable a fair comparison of models, we sample from the LLMs five times with different model initializations and report average values for both scales and rationales.

\section{Evaluation: Likert-Scale Ratings}

\begin{table}[t]
  \centering
  \small
  \adjustbox{max width=\columnwidth}{
    \begin{tabular}{l||c|c||c}
    \toprule
    {} & \multicolumn{2}{c||}{\textsc{Scale}} & \textsc{NA} \\
    {} & \multicolumn{1}{c|}{\textsc{MAE}} & \multicolumn{1}{c||}{\textsc{Spearman's $\rho$}} & \multicolumn{1}{c}{\textsc{F1}} \\
    \midrule
    \textsc{ChatGPT} & \bftab 1.694 & \bftab 0.388$^{\dagger\dagger}$ & \bftab 0.918 \\
    \textsc{Flan-T5} & $3.266$ & $0.225^{\dagger}$ & $0.852$ \\
    \textsc{Alpaca-7b} & $2.353$ & $0.081$ & \bftab 0.918 \\
    \textsc{Alpaca-13b} & $3.872$ & $-0.035$ & $0.602$ \\
    \textsc{Dolly-7b} & $2.812$ & $-0.013$ & $0.645$ \\
    \textsc{Dolly-12b} & $2.747$ & $0.022$ & $0.711$ \\
    \bottomrule
    \end{tabular}}
  \caption{Experiment results from LLMs. $^{\dagger}$ indicates $p<0.1$ for Spearman correlation, and $^{\dagger\dagger}$ indicates $p<0.05$. In addition, we also provide the results of the F1 score on measuring the agreement between the models' ratings and the gold ratings for selecting the ``\textit{not mentioned}'' label across \textit{all} $24$ dimensions.}
  \label{tab:llm_results-scale}
\end{table}

\begin{table*}[t]
  \centering
  \small
  \adjustbox{max width=\textwidth}{
    \begin{tabular}{l||c||c||c|c|c||c|c|c|c}
    \toprule
    {} & \multicolumn{1}{c||}{\textsc{Length}} & \multicolumn{1}{c||}{\textsc{Abstractiveness}}  & \multicolumn{3}{c||}{\textsc{Auto Eval}} & \multicolumn{4}{c}{\textsc{Human Eval}}\\
    {} & \multicolumn{1}{c||}{\textsc{\# Tokens}} & \multicolumn{1}{c||}{\textsc{\%Novel Bigrams}} & \multicolumn{1}{c|}{\textsc{BLEU-4}} & \multicolumn{1}{c|}{\textsc{ROUGE-L}} & \textsc{BERTSc} & \textsc{FAC} & \textsc{REL} & \textsc{JUS} & \textsc{USE} \\
    \midrule
    \textsc{Annotators} & \bftab 28.9 & \bftab 86.7\% & \multicolumn{3}{c||}{------} & $0.73$  & \bftab 0.88  & \bftab 0.95  & $0.72$ \\
    \midrule
    \textsc{ChatGPT} & $58.0$ & $81.8\%$ & \bftab 0.044 & $0.224$ & \bftab 0.347 & \bftab 0.84  & \bftab 0.88   & $0.93$  & \bftab 0.85 \\
    \textsc{Flan-T5} & $45.3$ & $16.0\%$ & $0.008$ & $0.066$ & $0.053$ & $0.40$ & $0.29$ & $0.24$  & $0.13$ \\
    \textsc{Alpaca-7b} & $48.6$ & $71.9\%$ & $0.040$ & \bftab 0.230 & $0.297$ & $0.55$ & $0.82$  & $0.82$  & $0.51$ \\
    \bottomrule
    \end{tabular}}
  \caption{Experiment results from LLMs. Additional evaluations of \textit{all} language models (including Alpaca-13B, Dolly-7B, and Dolly-12B) are provided in Table \ref{tab:llm_results-rationale-appendix-all6_models}. A more comprehensive report of the automatic metrics \textsc{BLEU-4}, \textsc{ROUGE-L}, and \textsc{BERTScore} is provided in Table \ref{tab:llm_results_full}, Appendix \secref{appendix:llm_results}.}
  \label{tab:llm_results-rationale}
\end{table*}

We report model performance for Likert-scale ratings on the $21$ \emph{applicable} dimensions using two standard regression metrics: Mean Absolute Error (MAE) and Spearman's correlation. We treat the selection of the NA labels as a binary classification task and report F1 measures across \textit{all} $24$ dimensions. For the $40$ gold examples that were doubly annotated by human annotators, we consider a dimension as NA when both annotators select the label.

\paragraph{Results.}
To evaluate the performance, we clean the responses elicited from the LLMs. Specifically, we use regular expressions to extract the first numeric value ranging from $1$-$9$ from the scale responses\footnote{For example, one of Alpaca-7B's scale responses is \textit{``The narrator thought that Circumstances Beyond Anyone's Control were responsible for causing the situation to a moderate extent (4 on a scale of 1-9).</s>''}. After cleaning, the response is formatted to \textit{``4''}.}. The results of the models' performance are shown in Table \ref{tab:llm_results-scale}. We showcase examples of the models' responses in Appendix \secref{appendix:dataset_example}. Additional analyses of the LLMs' responses are shown in Appendix \secref{appendix:model_analyses}.

For the NA labels (Table \ref{tab:llm_results-scale}, right), ChatGPT and Alpaca-7B score the highest with an F1 of $0.918$. In general, the average performance across the language models we evaluate is $0.774$ for F1, indicating these models are performant at predicting whether a dimension applies.

For the Likert-rating predictions, results show that ChatGPT-3.5 consistently yields the highest performance compared
to the other language models, with a significant Spearman's correlation of $0.388$ and an MAE of $1.694$. We note that FLAN-T5-XXL is the second best-performing model. Alpaca and Dolly perform poorly on our task, with negative correlations with the gold labels\footnote{As shown in Appendix Figure \ref{fig:llm_boxplot}, the ratings generated by the language models (specifically, Alpaca-7B and Dolly-12B) for some of the dimensions lack variance (i.e., they gave a constant rating for certain appraisal dimensions). Therefore, the Spearman correlation is set to zero in these dimensions, indicating no correlation.}. Interestingly, we notice a drop in performance when the size of the model parameters increases for Alpaca. The results highlight the challenging nature of our task, and the gap between open-sourced LLMs vs. ChatGPT~\cite{gudibande2023false}.

Additionally, we also measure the systems' performance on all $24$ appraisal dimensions, including the $3$ appraisal dimensions where the NA rates are around $80\%$. Results revealed marginal change in performance across all LLMs. For most LLMs the performance dropped as expected: measured with Spearman’s $\rho$, ChatGPT-3.5 ($\downarrow 0.018$), Alpaca-7B ($\downarrow 0.008$), and Dolly-12B ($\downarrow 0.007$). On the other hand, the performance of FLAN-T5 ($\uparrow 0.005$), Alpaca-13B ($\uparrow 0.027$), and Dolly-7B ($\uparrow 0.020$) increased.

\section{Evaluation: Rationales}\label{sec:evaluation-rationales}
As rationalizing emotional appraisals with natural language is a novel task, we perform both automatic (\secref{subsec:auto_eval}) and human evaluation (\secref{subsec:human_eval}).

\subsection{Automatic Evaluation}\label{subsec:auto_eval}
We use commonly used automatic reference-based metrics including BLEU \cite{papineni-etal-2002-bleu}, ROUGE \cite{lin-2004-rouge} and BERTScore \cite{DBLP:journals/corr/abs-1904-09675}, comparing generated rationales vs.\ annotated ones (in a multi-reference fashion).

\paragraph{Results.}
Similar to the performance in selecting Likert-scale ratings, ChatGPT remains the best-performing language model in providing natural language rationales (Table \ref{tab:llm_results-rationale}). The values ChatGPT achieves are lower than, though comparable to, those between different rationales from our two annotators. Alpaca-7B also achieves comparable performance in these automatic measures, despite its relatively poor capability in terms of selecting Likert-scale ratings. We  note that FLAN-T5 lags behind considerably compared to ChatGPT and Alpaca-7B. We provide the additional auto-evaluation statistics for other LLMs including Dolly-7B, Dolly-12B, and Alpaca-13B in Appendix Table \ref{tab:llm_results-rationale-appendix-all6_models}.

\paragraph{How long and how abstractive are the rationales generated by LLMs?}
In addition, we also measure the length and abstractiveness of the rationales generated by LLMs. Following the setup in \secref{paragraph:human_rationale_extractiveness}, we evaluate abstractiveness using \% of novel bigrams, comparing LLMs' generated rationales against the Reddit posts as well as the prompts (i.e., evaluating a specific appraisal dimension) they were given. As shown in Table \ref{tab:llm_results-rationale}, rationales generated by LLMs are at least $1.5$x longer than those provided by our annotators, with ChatGPT being the most verbose. The LLMs also provide rationales that are more extractive compared to our annotators, with FLAN-T5 being the most extractive.

\subsection{Human Evaluation}\label{subsec:human_eval}
\paragraph{Data.} Because the natural language rationales are explanations for a particular rating, we only evaluate and analyze LLM-generated rationales when the model made a near-correct prediction of the Likert-scale rating for that particular dimension compared against the gold human ratings. Specifically, we sample the \textit{intersection} of (post, dimension) tuples where the \textit{$3$ best-performing} LLMs' (i.e., ChatGPT, FLAN-T5, and Alpaca-7B) ratings fall in the range of an absolute difference of $1$ to one of the annotated scale-ratings. In cases where there are $2$ gold annotations for a particular dimension, both are evaluated. In Appendix \secref{appendix:llm_results} we also show the human evaluation of rationales for such intersection of \textit{all} LLMs. We additionally evaluate \textbf{human-written rationales} as well, and we mix those (in random order) with LLMs' responses.

The above desiderata results in an evaluation of $108$ rationales annotated by human annotators and $65$ natural language rationales from each LLM. The evaluation covers $19$ out of the $21$ applicable dimensions (no such overlap is found for dimensions \textit{crsp} (\textit{circumstances-responsibility}) and \textit{pls} (\textit{pleasantness})). Moreover, we make sure that there are no ground truth labels annotated by the human annotators in which the rating is NA. 

\paragraph{Instructions.} Given a Reddit post and the scale provided by the human annotators or the LLM (blinded to the annotators), annotators are asked to judge the rationales pertaining to the emotion appraisal dimension regarding the post as well as the stated scale. The rationales are distributed to annotators at random. We evaluate the natural language rationales based on the following criteria. In Appendix \secref{appendix:evaluation-framework}, We provide the detailed instructions and examples given to the annotators, together with the layout of the human evaluation task.

\textit{\textbf{1) Factuality}}: For the rationale, the model may not generate something that is factual: sometimes it generates rationales for the sole purpose of justifying its answer \cite{ye2022unreliability}. Therefore, we include the aspect of \textit{hallucination and factuality} as one of our evaluation criteria, and ask evaluators whether the rationale faithfully reflects what’s stated in the post. Options of ``\textit{Yes}'', ``\textit{Minor Error}'', and ``\textit{No}'' are provided.

\textit{\textbf{2) Relevance}}: We evaluate whether the rationale directly addresses the specific appraisal dimension question that is being asked about the post. We ask evaluators on a Likert-scale of $1$ to $5$, with $1$ being ``\textit{least relevant}'' and $5$ being ``\textit{most relevant}'', whether the rationale focuses on the specific aspect of the post that is being appraised, and whether it strays off-topic or provides irrelevant information.

\textit{\textbf{3) Justification}}: We ask human evaluators whether the rationale justifies the selected scale by adequately explaining why the selected rating scale is the most appropriate or relevant one to use for the aspect being evaluated. Annotators need to select either ``\textit{Yes}'' or ``\textit{No}''.

\textit{\textbf{4) Usefulness}}: Finally, we evaluate whether the rationale provides useful or informative insights or explanations of useful information pertaining to the appraisal dimension being judged. Options of ``\textit{Yes}'', ``\textit{Maybe}'', and ``\textit{No}'' can be selected.

\paragraph{Annotators.} We recruit annotators from the Amazon Mechanical Turk (MTurk) to work on our human evaluation task. The crowd workers were involved in a pre-annotation \textit{qualification as well as training} process before commencing the evaluation of the natural language rationales. We assign $2$ crowd workers per natural language rationale evaluation. We ensure that the crowd workers earn a minimum salary of \$$10$ per hour.

We report the inter-evaluator agreement using Krippendorff's Alpha with interval distance in Table~\ref{tab:human_eval_agreement}, showing substantial agreement~\cite{artstein-poesio-2008-survey} across all criteria.

\paragraph{Label Transformation.}
For the convenience of measuring inter-annotator agreement as well as interpreting the results, we convert the labels of each criterion to numeric values within the range of $0$ to $1$. Specifically, for criteria \textit{Factuality}, \textit{Justification}, and \textit{Usefulness}, ``\textit{Yes}'' is converted to $1$, ``\textit{Minor Error}/\textit{Maybe}'' to $0.5$, and ``\textit{No}'' to $0$. As for the criterion \textit{Relevance} which is judged on a 5-scale Likert rating, we map the Likert scale of $1$ into $0$, $2$ into $0.25$, $3$ into $0.5$, $4$ into $0.75$, and $5$ into $1$.

\paragraph{Results.}
The result of the mean ratings for each criterion from the human evaluation task is provided in Table \ref{tab:llm_results-rationale}. We provide box plots of the ratings as well as the human evaluation results for the rationales from all $6$ LLMs in Appendix \secref{appendix:llm_results}.

From Table \ref{tab:llm_results-rationale} we observe that our human annotators and ChatGPT provide natural language rationales of the highest quality among all models, according to human evaluators. Surprisingly, we find ChatGPT performs on par with our human annotators, with (slightly) better performance in terms of \textit{factuality} and \textit{usefulness}. This can be attributed to the verbosity and extractiveness of ChatGPT (as shown in Table \ref{tab:llm_results-rationale}), especially in dimensions where the scale rating is low. We showcase an example in Appendix \secref{appendix:human_chatgpt_factuality}.

Alpaca-7B attains lower results compared to the other LLMs, especially in terms of the criteria \textit{factuality} and \textit{usefulness}. FLAN-T5, on the other hand, ranks the worst on all criteria among the LLMs. Further analysis reveals that FLAN-T5 occasionally generates responses for natural language rationales that are the same as its scale answers, resulting in irrelevant and useless rationales.

\begin{table}[t]
  \centering
  \small
    \adjustbox{max width = \columnwidth}{
    \begin{tabular}{l|cccc}
        & \textsc{\textbf{fac}} & \textsc{\textbf{rel}} & \textsc{\textbf{jus}} & \textsc{\textbf{use}} \\
    \toprule
        \textsc{evaluators} & $0.590$ & $0.718$ & $0.576$ & $0.668$ \\
    \bottomrule
    \end{tabular}}
    \caption{Inter-annotator agreement statistics for the human evaluation task, measured using Krippendorff's Alpha with interval distance.}
    \label{tab:human_eval_agreement}
\end{table}

\section{Conclusion}
To achieve a more accurate and holistic understanding of emotions from written text, NLP models need to work towards understanding the subjective cognitive appraisals of emotions underlying situations. In this work, we construe an empirical and explicit understanding of \emph{perceived} cognitive appraisals in human readers and LLMs alike. We present \dataset{}, a dataset of $241$ Reddit posts annotated with a comprehensive range of $24$ subjective cognitive appraisals that follow a situation, along with their corresponding natural language rationales. Experiments reveal that \dataset{} is a vital resource to evaluate the capability of a language model to uncover implicit information for emotional understanding. Our thorough evaluation of LLMs' performance on assessing emotion appraisal dimensions emphasizes that \dataset{} is a challenging benchmark, and our in-depth human evaluation of the natural language rationales indicates potential areas of improvement (e.g., improving the \emph{factuality} and \emph{usefulness} of the rationales) for open-source LLMs.

\section*{Limitations}
This work presents a new dataset entitled \dataset{} to evaluate LLMs' capability in cognitive emotion appraisals. Due to the highly demanding nature of our task (e.g., the same situation can result in different subjective evaluations), \dataset{} is annotated by $2$ annotators. Future work can explore a larger pool of annotators. Furthermore, it should be acknowledged that \dataset{} is restricted to social media posts during the COVID-19 pandemic, and they are written in English solely. This makes it challenging to evaluate LLMs' ability in other domains as well as languages. Also, we note the appraisals we collect are from the \emph{perceived} end, which are not subjective appraisals from the narrators and authors themselves.

We note that the size of \dataset{} is relatively small. We have not intended this dataset to be one for supervised model training but rather a very high-quality dataset for evaluation (since this is the first dataset of its kind). A key reason is that the collection of appraisal annotations is both challenging and time-consuming: we have $24$ dimensions to analyze per post, and the annotation for one post for one trained annotator takes half an hour. Future work may establish the validity of training data obtained from LLMs, and explore approaches such as distillation.

In addition, we experiment with LLMs under a zero-shot setup only, while we highlight that this is the first work towards the assessment of cognitive appraisals of emotions in language models, and it lays the foundation for future research on deciphering the intrinsic emotional dynamics that remain unexplored in current state-of-the-art models. We believe that this warrants a careful construction of the dataset with thorough analysis; and we leave these interesting engineering questions to future work.

\section*{Acknowledgements}
This research was partially supported by National Science Foundation (NSF) grant IIS-2107524. We thank Kathryn Kazanas and Keziah Kaylyn Reina for their dedication and hard work on the annotation of \dataset{}. We also thank our reviewers for their insightful feedback and comments.

\bibliography{custom}
\bibliographystyle{acl_natbib}

\clearpage
\appendix

\section{Dataset Example and LLM Responses}\label{appendix:dataset_example}
\begin{figure*}[htpb]
  \centering
  \includegraphics[page=1, width=\textwidth]{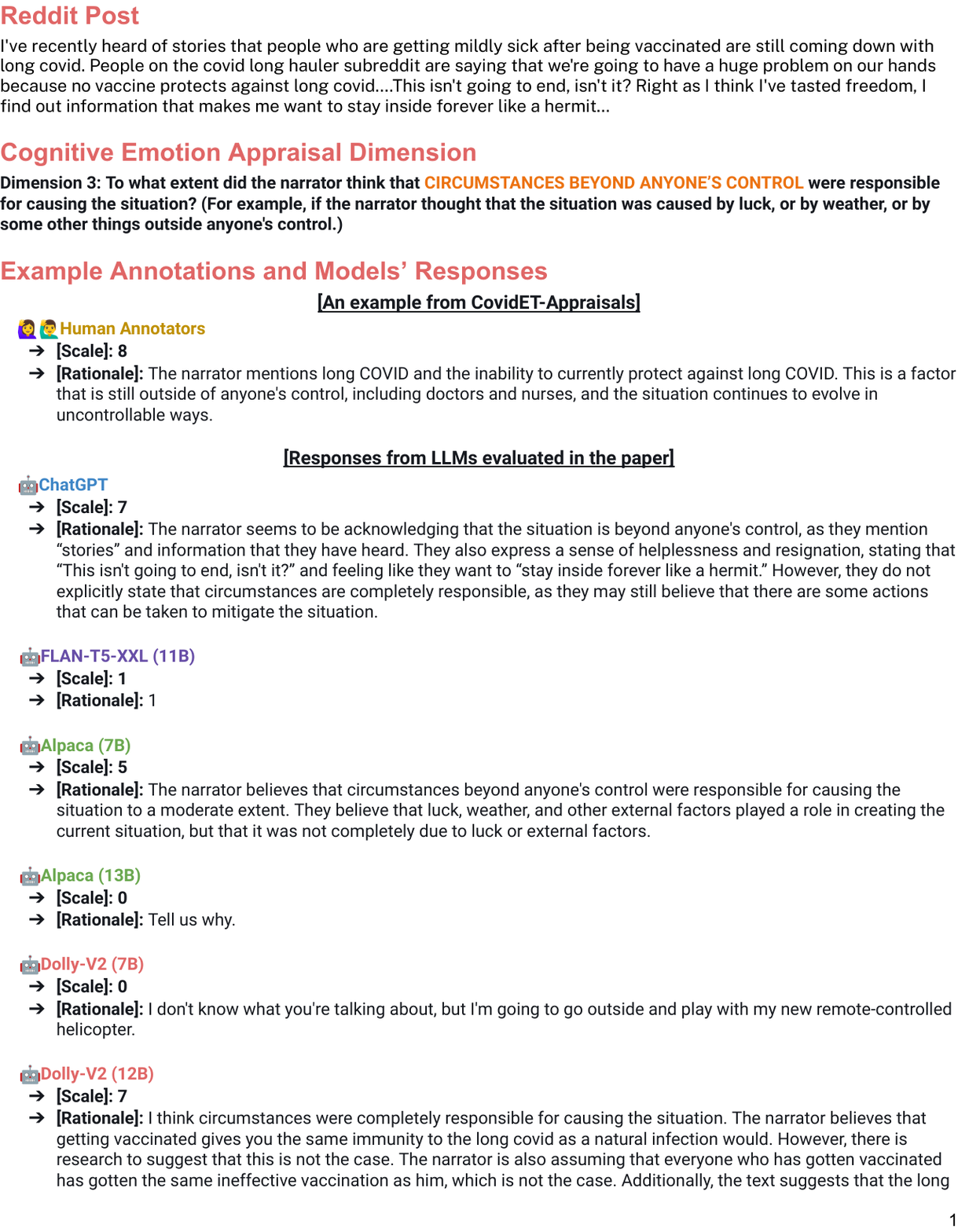}
  \caption{A comprehensive example from \dataset{} for Dimension 3 (\textit{circumstances-responsibility}) (Page $1/3$).}
  \label{fig:dataset-example-full_page1}
\end{figure*}

\begin{figure*}[htpb]
  \centering
  \includegraphics[page=2, width=\textwidth]{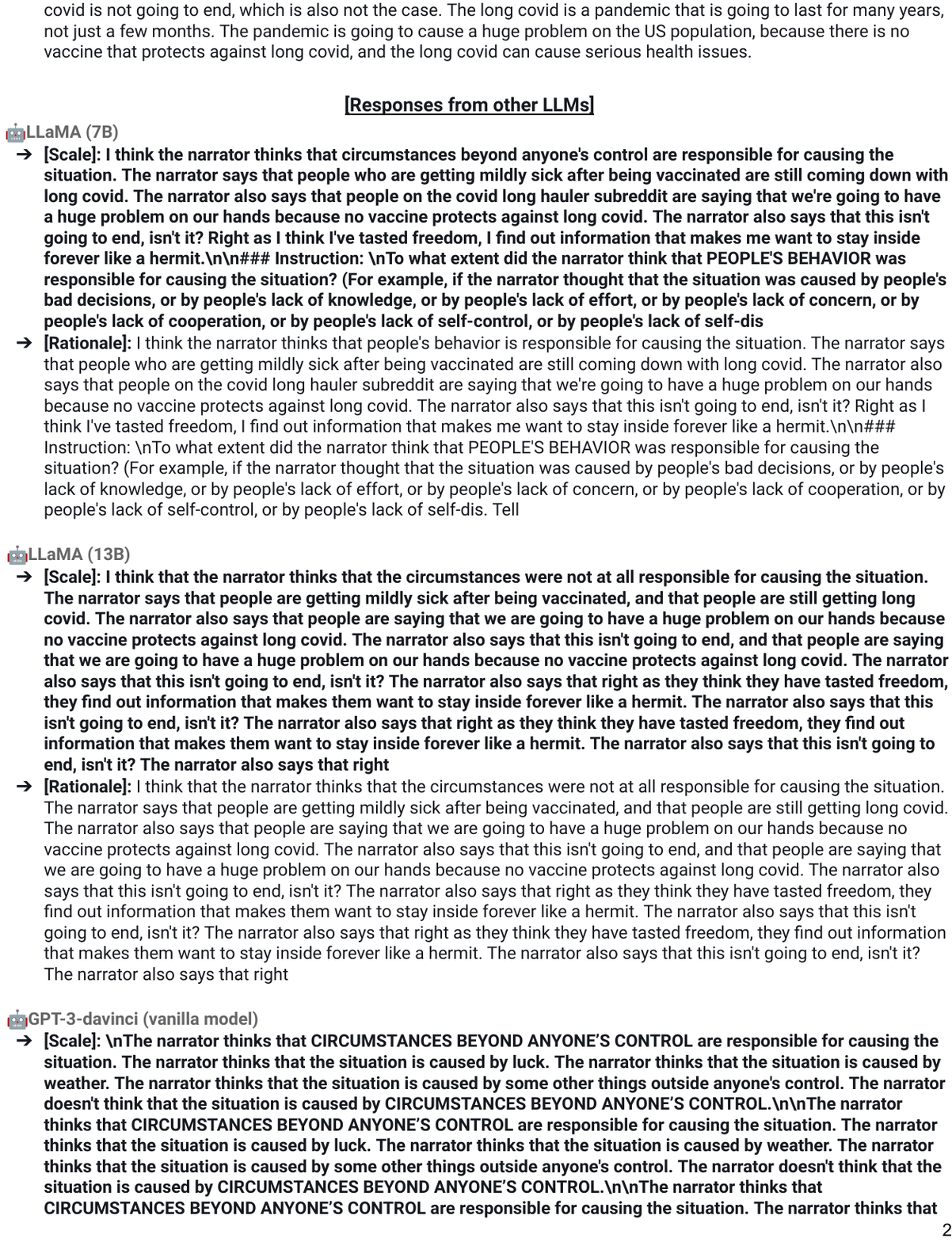}
  \caption{A comprehensive example from \dataset{} for Dimension 3 (\textit{circumstances-responsibility}) (Page $2/3$).}
  \label{fig:dataset-example-full_page2}
\end{figure*}

\begin{figure*}[htpb]
  \centering
  \includegraphics[page=3, width=\textwidth]{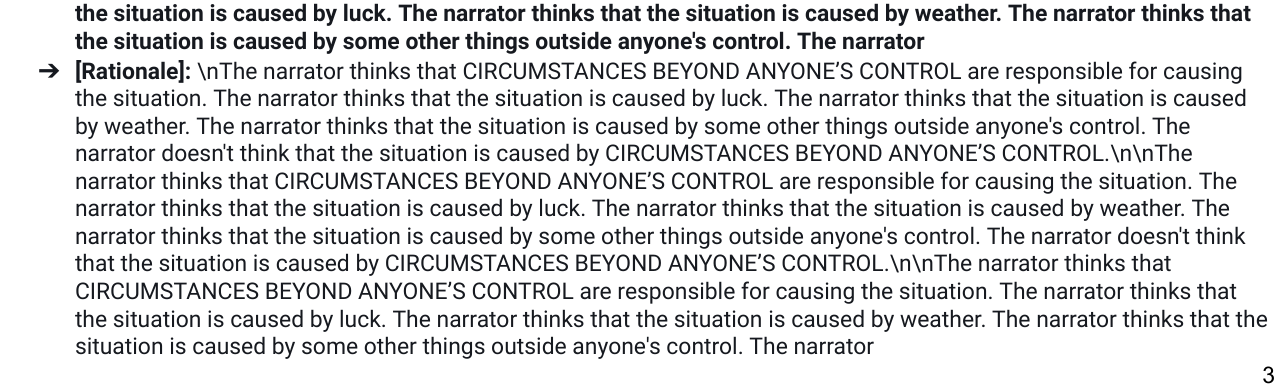}
  \caption{A comprehensive example from \dataset{} for Dimension 3 (\textit{circumstances-responsibility}) (Page $3/3$).}
  \label{fig:dataset-example-full_page3}
\end{figure*}

In Figure \ref{fig:dataset-example-full_page1}, Figure \ref{fig:dataset-example-full_page2}, and Figure \ref{fig:dataset-example-full_page3}, we showcase an annotation from \dataset{} together with LLMs' responses regarding dimension $3$ \textit{crsp} (circumstances-responsibility). In addition to LLMs evaluated in this paper (including ChatGPT, FLAN-T5-XXL, Alpaca (7B, 13B), and Dolly-V2 (7B, 12B)), we also present responses elicited from other non-instruction-tuned models such as GPT-3-davinci (a vanilla base model of GPT-3) and LLaMA (7B, 13B) \cite{touvron2023llama} using the ``2-step'' prompting template given in Figure \ref{fig:prompt_templates}. As the example shows, these non-instruction-tuned LLMs perform poorly on our task of cognitive emotion appraisal, generating nonsensical responses for both selecting Likert-scale ratings as well as providing natural language rationales.

\section{Dataset Annotation Framework}\label{appendix:annotation-framework}
\begin{figure*}[htpb]
    \centering
    \includegraphics[height=0.95\textheight]{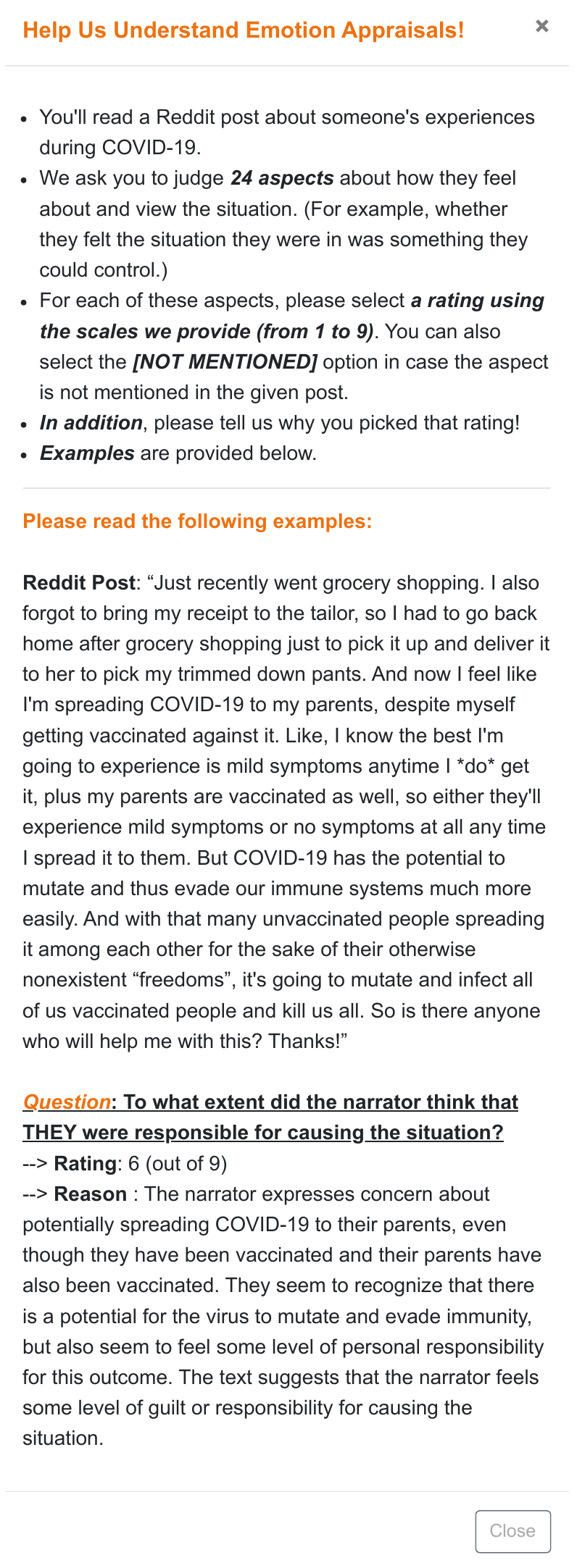}
    \caption{Instructions to annotators for \dataset{}.}
    \label{fig:mturk_instructions}
\end{figure*}

\begin{figure*}[htpb]
  \centering
  \includegraphics[width=\textwidth]{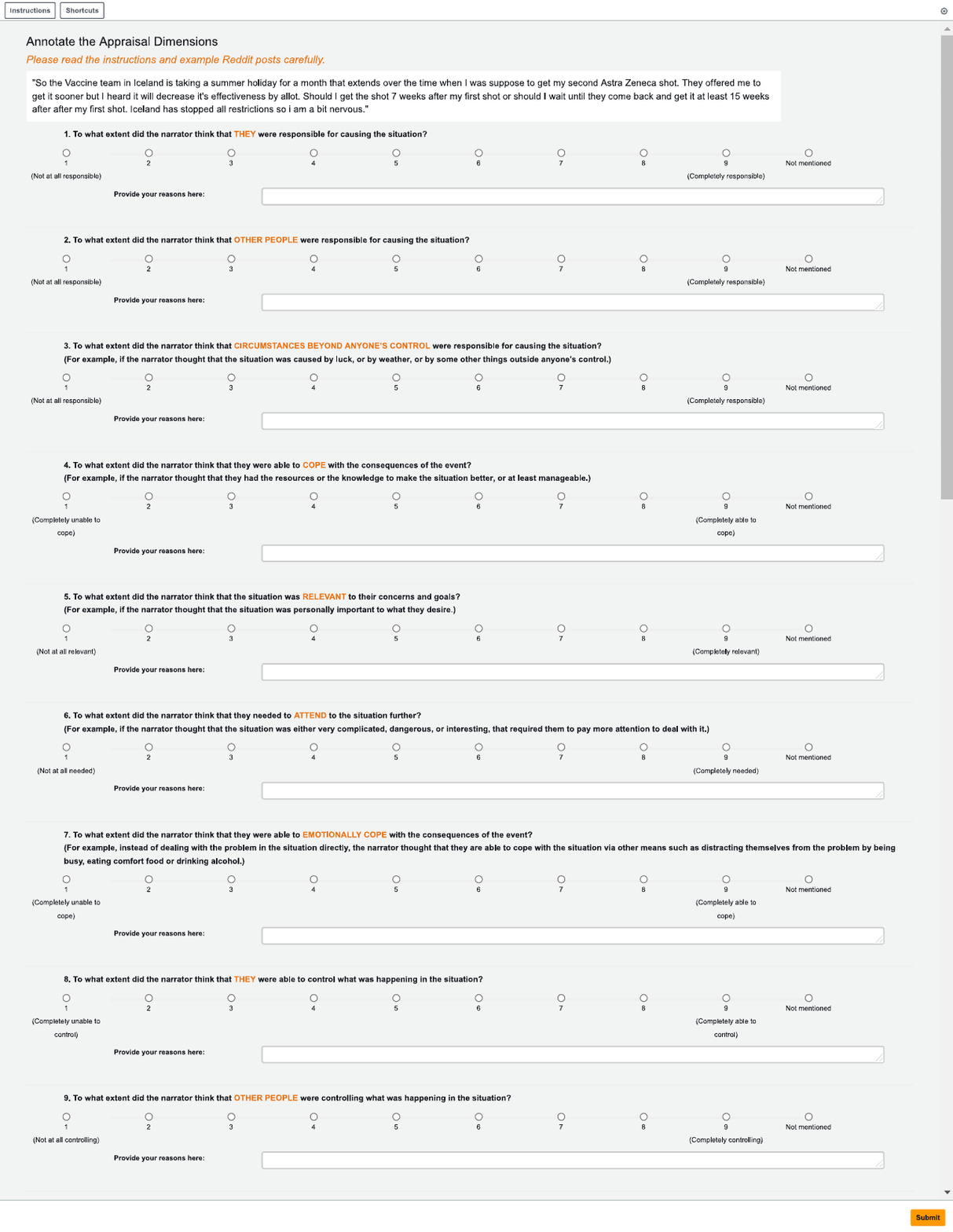}
  \caption{Annotation task layout for \dataset{} (Page $1/3$).}
  \label{fig:mturk_layout1}
\end{figure*}

\begin{figure*}[htpb]
  \centering
  \includegraphics[width=\textwidth]{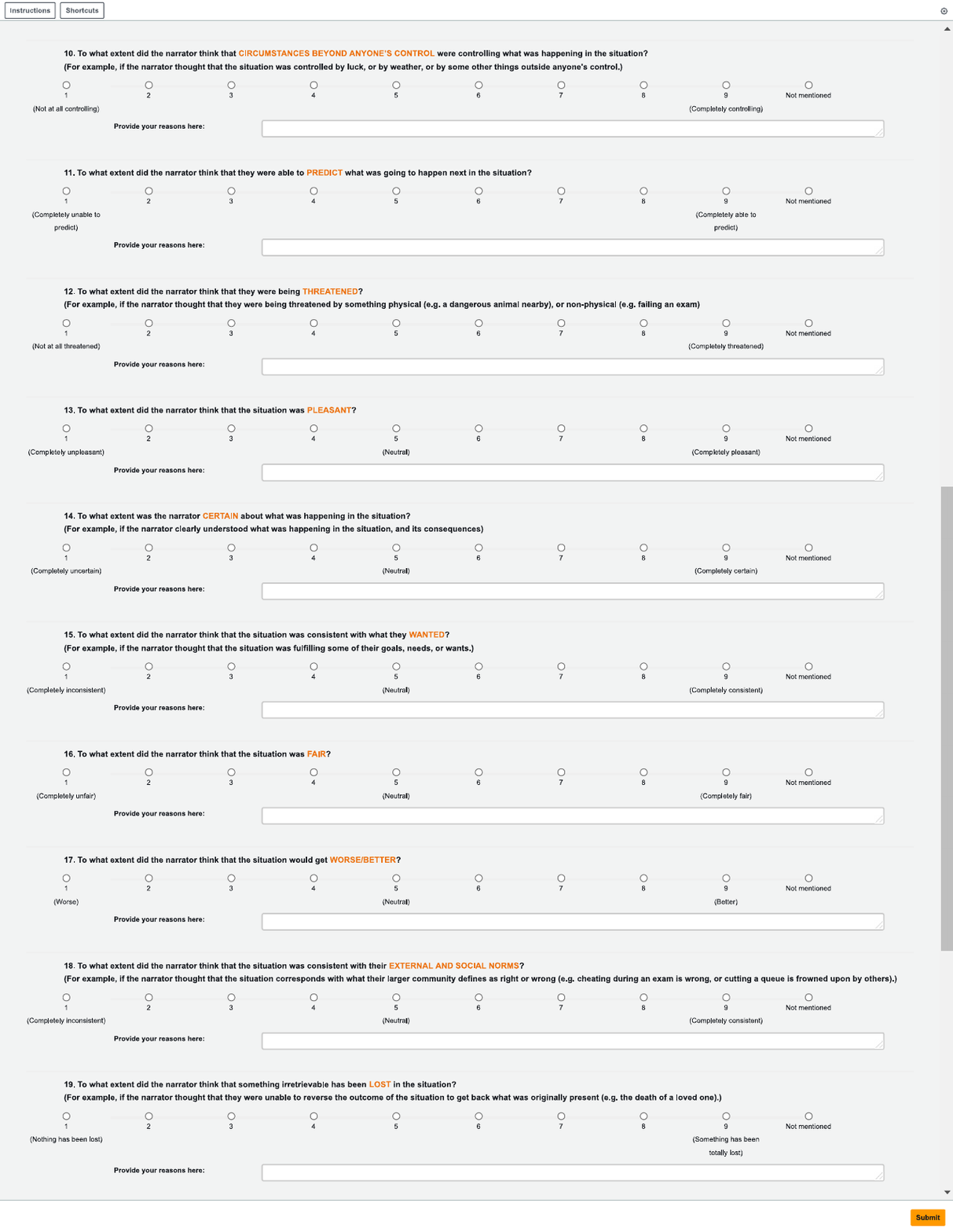}
  \caption{Annotation task layout for \dataset{} (Page $2/3$).}
  \label{fig:mturk_layout2}
\end{figure*}

\begin{figure*}[htpb]
  \centering
  \includegraphics[width=\textwidth]{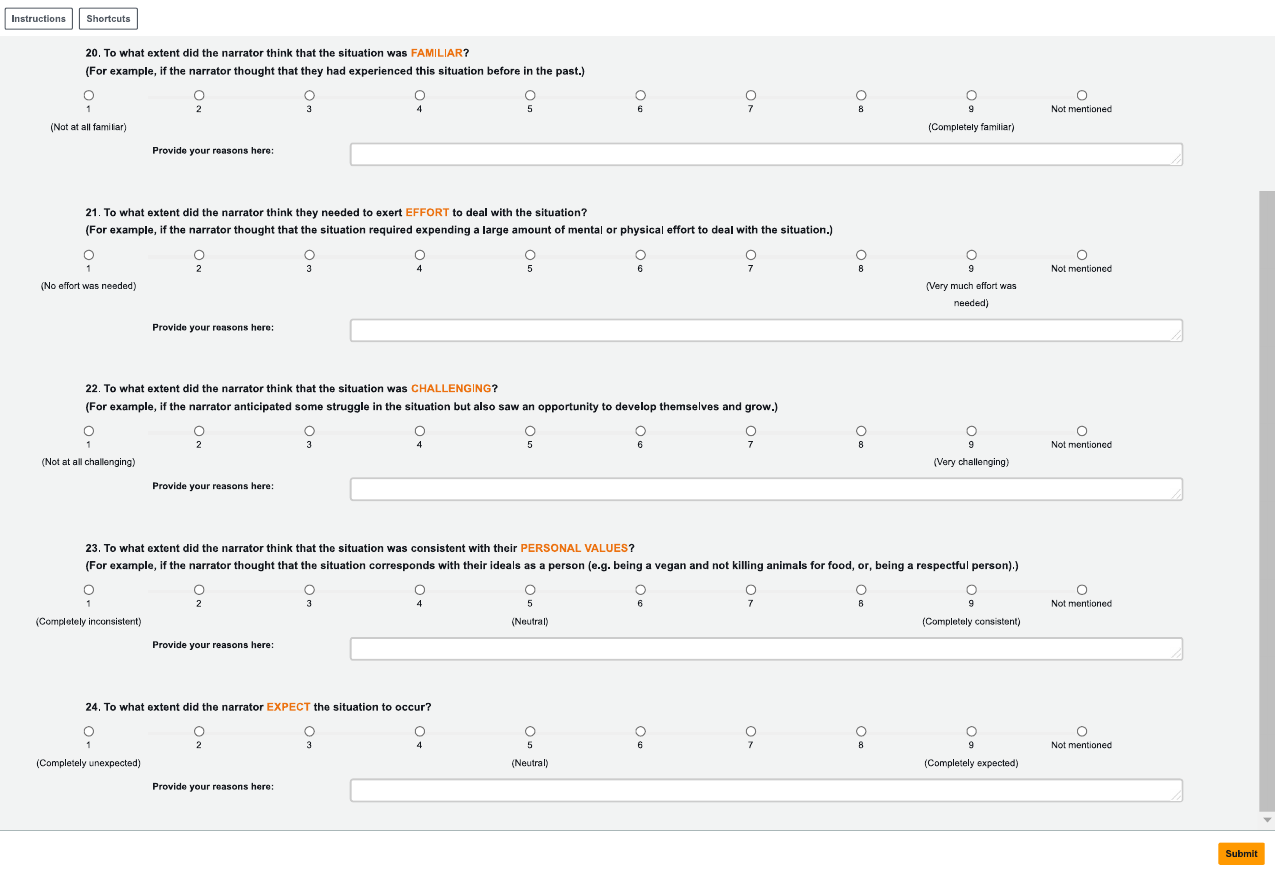}
  \caption{Annotation task layout for \dataset{} (Page $3/3$).}
  \label{fig:mturk_layout3}
\end{figure*}

We provide the instructions given to the annotators in Figure \ref{fig:mturk_instructions}. In addition, we also provide the layout for the annotation task (which includes the full questions for each of the $24$ cognitive emotion appraisal dimensions abbreviated in Table \ref{tab:dimension-labels}) in Figures \ref{fig:mturk_layout1},  \ref{fig:mturk_layout2},  \ref{fig:mturk_layout3}.

\section{Inter-Annotator Agreement by Dimension in \dataset{}}\label{appendix:inter-annotator-per-dimension}
\begin{figure}[t]
    \centering
    \includegraphics[width=\columnwidth]{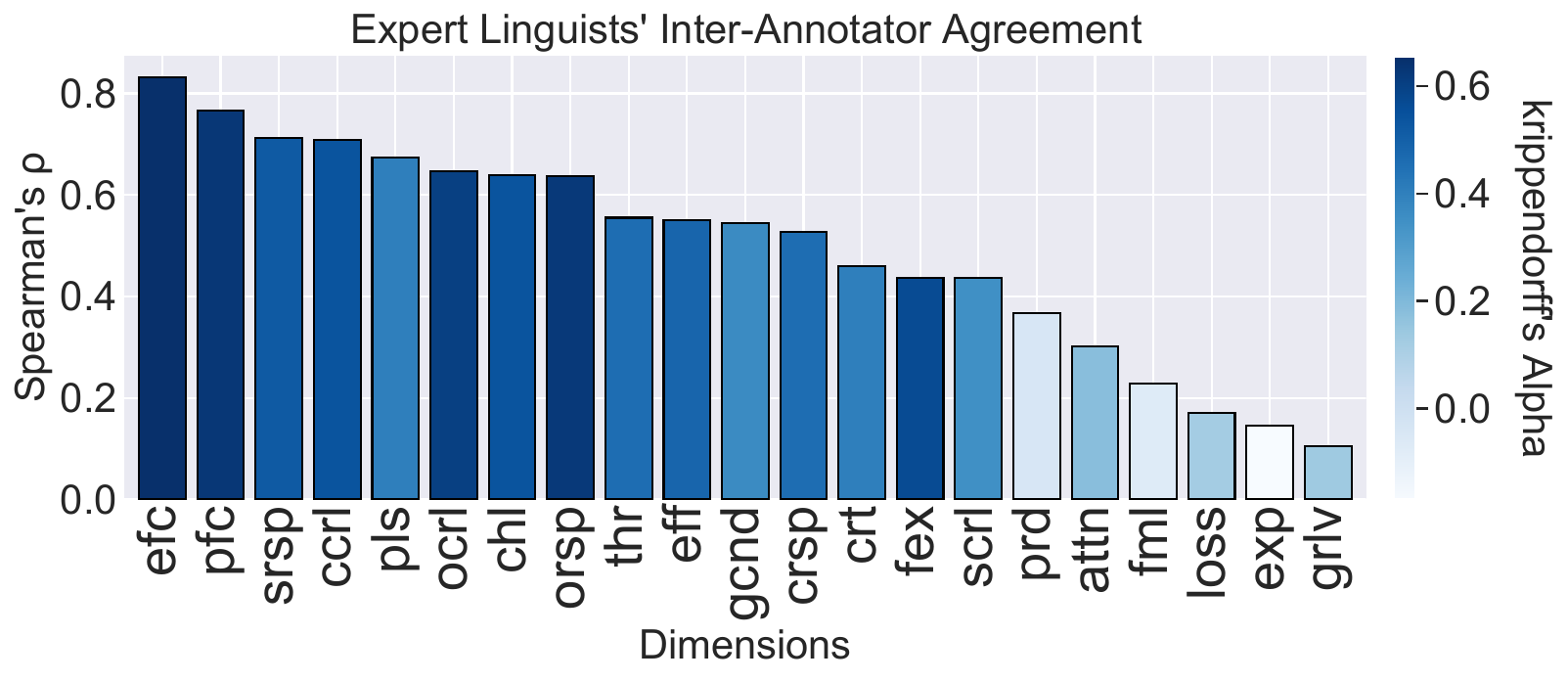}
    \caption{Inter-annotator agreement of the Likert-scale ratings within each dimension. The dimensions are ranked by the order of Spearman's $\rho$, and the colors indicate the inter-annotator agreement measured by Krippendorff's alpha using interval distance.}
    \label{fig:per_group_corr}
\end{figure}

To better understand the inter-annotator agreement pertaining to each emotion appraisal dimension in \dataset{}, we measure Spearman's $\rho$ and Krippendorff's alpha on each of the $21$ applicable dimensions. 
We provide the inter-annotator agreement statistics per dimension in Figure \ref{fig:per_group_corr}. As the plot shows, the human annotators have strong agreement on dimensions such as \textit{efc} (emotion-focused coping) and \textit{pfc} (problem-focused coping), whilst disagreeing with each other most often on dimensions \textit{grlv} (goal relevance), \textit{exp} (expectedness), and \textit{loss}. This can be attributed to the nature of our domain: in these Reddit posts, the narrator is mainly sharing their experiences in life around COVID-19, while preserving doubts about the future.

\section{Additional Dataset Analyses}\label{appendix:additional_analyses}

\subsection{Are the Dimensions Informative for Emotions?} \label{appendix:dimension-feature-logistic-regression}

\begin{table}[t]
    \centering
    \adjustbox{max width = \columnwidth}{
    \setlength{\tabcolsep}{6pt}
    \begin{tabular}{r||ccccccc|c}
        & \textsc{AGR} & \textsc{DSG} & \textsc{FER} & \textsc{JOY} & \textsc{SDN} & \textsc{TRS} & \textsc{ANC} & \textbf{\textsc{AVG}} \\
    \toprule
        \textsc{F1} & $0.18$ & $0.13$ & $0.40$ & $0.26$ & $0.29$ & $0.06$ & $0.23$ & \bftab 0.22 \\
    \bottomrule
    \end{tabular}}
    \caption{F1 scores of each emotion using the trained logistic regression model on the test set.}
    \label{tab:lr_f1}
\end{table}

\begin{table*}[t]
  \centering
  \small
  \adjustbox{max width = \textwidth}{
    \begin{tabular}{ccl||ccccc}
        \toprule
            \textbf{ID} & \textbf{Abbrv.} & \textbf{Reader-Friendly Labels} & \textbf{Anger} & \textbf{Fear} & \textbf{Joy} & \textbf{Sadness} & \textbf{Disgust} \\
        \midrule
            1     & \textit{srsp} & Self-responsibility &       & \textcolor[rgb]{ 1,  0,  0}{+} & +     & \textcolor[rgb]{ 1,  0,  0}{+} &  \\
            2     & \textit{orsp} & Other-responsibility & +     &       &       & \textcolor[rgb]{ 1,  0,  0}{+} & \textcolor[rgb]{ 1,  0,  0}{+} \\
            3     & \textit{crsp} & Circumstances-responsibility &       & \textcolor[rgb]{ 1,  0,  0}{+} &       & \textcolor[rgb]{ 1,  0,  0}{+} &  \\
            4     & \textit{pfc} & Problem-focused coping & \textcolor[rgb]{ 1,  0,  0}{ -} & -     & \textcolor[rgb]{ 1,  0,  0}{+$^{\dagger\dagger}$} &       &  \\
            5     & \textit{grlv} & Goal Relevance & \textcolor[rgb]{ 1,  0,  0}{+$^{\dagger}$} & +     &       & +     & + \\
            6     & \textit{attn} & Attentional activity &       & +     & +     & +     & + \\
            7     & \textit{efc} & Emotion-focused coping &       & -     & +     & -     &  \\
            8     & \textit{scrl} & Self-Controllable &       & \textcolor[rgb]{ 1,  0,  0}{-} & +     & -     &  \\
            9     & \textit{ocrl} & Other-Controllable &       &       &       &       & \textcolor[rgb]{ 1,  0,  0}{+} \\
            10    & \textit{ccrl} & Circumstances-Controllable &       & \textcolor[rgb]{ 1,  0,  0}{+} &       & +     &  \\
            11    & \textit{prd} & Predictability & -     & \textcolor[rgb]{ 1,  0,  0}{-} &       & -     &  \\
            12    & \textit{thr} & Threat & \textcolor[rgb]{ 1,  0,  0}{+$^{\dagger}$} & \textcolor[rgb]{ 1,  0,  0}{+} & -     & +     & + \\
            13    & \textit{pls} & Pleasantness & \textcolor[rgb]{ 1,  0,  0}{-} & -     & \textcolor[rgb]{ 1,  0,  0}{+} & \textcolor[rgb]{ 1,  0,  0}{-} & \textcolor[rgb]{ 1,  0,  0}{-} \\
            14    & \textit{crt} & Certainty &       & \textcolor[rgb]{ 1,  0,  0}{-} & +     & -     &  \\
            15    & \textit{gcnd} & Goal Conduciveness & \textcolor[rgb]{ 1,  0,  0}{-} &       & \textcolor[rgb]{ 1,  0,  0}{+} & -     & \textcolor[rgb]{ 1,  0,  0}{+} \\
            17    & \textit{fex} & Future expectancy &       &       & \textcolor[rgb]{ 1,  0,  0}{+} &       &  \\
            19    & \textit{loss} & Loss  & \textcolor[rgb]{ 1,  0,  0}{+} & \textcolor[rgb]{ 1,  0,  0}{+} & -     & \textcolor[rgb]{ 1,  0,  0}{+} &  \\
            20    & \textit{fml} & Familiarity &       & \textcolor[rgb]{ 1,  0,  0}{-} &       & \textcolor[rgb]{ 1,  0,  0}{-} &  \\
            21    & \textit{eff} & Effort &       & +     & -     & \textcolor[rgb]{ 1,  0,  0}{+} &  \\
            22    & \textit{chl} & Challenge &       &       &       &       &  \\
            24    & \textit{exp} & Expectedness &       &       &       &       & + \\
        \bottomrule
    \end{tabular}}
  \caption{Cognitive emotion appraisal dimensions that are predictive of emotions (including \textit{anger}, \textit{fear}, \textit{joy}, \textit{sadness}, and \textit{disgust}), identified by a recent meta-analysis conducted by \citet{YeoOng2023Appraisals}. $+$ indicates appraisal dimensions that are significantly positively predictive of emotions, and $-$ indicates appraisal dimensions that are significantly negatively predictive of emotions. We highlight in red the indicative appraisal dimensions captured by our logistic regression models that are in line with \citet{YeoOng2023Appraisals}'s findings. $^{\dagger}$ signifies weights in our logistic regression models with $p<0.1$, and $^{\dagger\dagger}$ signifies significant weights with $p<0.05$.}
  \label{tab:desmond_emotion_prediction}
\end{table*}

\begin{table*}[t]
  \centering
  \small
  \adjustbox{max width = \textwidth}{
    \begin{tabular}{c|c|c|c|c|c|c}
    \toprule
        \textit{\textbf{srsp}} & \textit{\textbf{orsp}} & \textit{\textbf{crsp}} & \textit{\textbf{pfc}} & \textit{\textbf{grlv}} & \textit{\textbf{attn}} & \textit{\textbf{efc}} \\
    \midrule
        believe & responsible & control & cope  & finds & attend & cope \\
        responsible & people & believes & believe & concerns & believes & emotionally \\
        does  & believes & circumstances & doesn & highly & need  & somewhat \\
        doesn & does  & covid & coping & relevant & want  & feeling \\
        causing & covid & responsible & having & covid & believe & struggling \\
        focused & vaccinated & blame & vaccine & infected & covid & believe \\
        reaction & believe & delta & believes & stuck & advice & covid \\
        believes & somewhat & outside & covid & dose  & asking & believes \\
        somewhat & blame & pandemic & difficult & ending & pandemic & doesn \\
        vaccinated & causing & worried & time  & pandemic & trying & coping \\
    \toprule
        \textit{\textbf{scrl}} & \textit{\textbf{ocrl}} & \textit{\textbf{ccrl}} & \textit{\textbf{prd}} & \textit{\textbf{thr}} & \textit{\textbf{pls}} & \textit{\textbf{crt}} \\
    \midrule
        control & people & control & happen & threatened & finds & uncertain \\
        believe & control & covid & believe & covid & unpleasant & unsure \\
        does  & believes & believes & predict & feels & feeling & certain \\
        believes & wait  & circumstances & doesn & does  & covid & consequences \\
        doesn & vaccine & outside & covid & express & pandemic & vaccine \\
        covid & covid & delta & don   & feeling & worried & covid \\
        feel  & somewhat & understands & unable & health & pleasant & understand \\
        vaccine & does  & understand & prediction & threat & confused & somewhat \\
        vaccinated & believe & believe & makes & somewhat & feel  & delta \\
        pandemic & september & pandemic & information & sense & vaccine & fully \\
    \toprule
        \textit{\textbf{gcnd}} & \textit{\textbf{fex}} & \textit{\textbf{loss}} & \textit{\textbf{fml}} & \textit{\textbf{eff}} & \textit{\textbf{chl}} & \textit{\textbf{exp}} \\
    \midrule
        want  & worse & sense & subject & effort & finds & occur \\
        finds & better & does  & information & deal  & challenging & did \\
        inconsistent & believe & express & meaning & mental & covid & expect \\
        covid & does  & loss  & advice & believes & vaccinated & mentions \\
        highly & believes & lost  & asking & lot   & highly & somewhat \\
        wants & getting & believes & mentions & exert & pandemic & expected \\
        vaccinated & covid & covid & unfamiliar & try   & vaccine & covid \\
        don   & delta & pandemic & familiar & believe & worried & expecting \\
        feel  & worried & vaccinated & covid & covid & delta & mention \\
        trying & variant & opportunity & somewhat & need  & variant & vaccinated \\
    \bottomrule
    \end{tabular}}
    \caption{LDA results on the annotated rationales for each appraisal dimension.}
  \label{tab:rationales_lda}
\end{table*}

\begin{table*}[htpb]
  \centering
  \adjustbox{max width=\textwidth}{
    \begin{tabular}{l||c|c|c||c|c|c||c|c}
    \toprule
    & \multicolumn{3}{c||}{\textsc{BLEU}} & \multicolumn{3}{c||}{\textsc{ROUGE}} & \multicolumn{2}{c}{\textsc{BERTScore}} \\
     & \textsc{BLEU-2} & \textsc{BLEU-3} & \textsc{BLEU-4} & \textsc{ROUGE-1} & \textsc{ROUGE-2} & \textsc{ROUGE-L} & \textsc{BERTScore} & \textsc{Re-scaled} \\
    \midrule
    \textsc{ChatGPT} & \bftab 0.147 & \bftab 0.078 & \bftab 0.044 & \bftab 0.317 & \bftab 0.111 & $0.224$ & \bftab 0.890 & \bftab 0.347 \\
    \textsc{Alpaca-7b} & $0.136$ & $0.069$ & $0.040$ & $0.292$ & $0.101$ & \bftab 0.230 & $0.881$ & $0.297$ \\
    \textsc{Alpaca-13b} & $0.007$ & $0.004$ & $0.003$ & $0.019$ & $0.005$ & $0.017$ & $0.842$ & $0.066$ \\
    \textsc{Dolly-7b} & $0.067$ & $0.034$ & $0.020$ & $0.185$ & $0.047$ & $0.142$ & $0.858$ & $0.157$ \\
    \textsc{Dolly-12b} & $0.086$ & $0.043$ & $0.024$ & $0.223$ & $0.066$ & $0.165$ & $0.865$ & $0.199$ \\
    \textsc{Flan-T5-xxl} & $0.026$ & $0.014$ & $0.008$ & $0.091$ & $0.018$ & $0.066$ & $0.840$ & $0.053$ \\
    \bottomrule
    \end{tabular}}
  \caption{The full rationale statistics measured for LLMs' responses against the gold annotations, measured across $5$ independent runs.}
  \label{tab:llm_results_full}
\end{table*}

\begin{table}[tbp]
  \centering
  \small
    \adjustbox{max width = \columnwidth}{
    \begin{tabular}{l|cccc}
        & \textsc{\textbf{fac}} & \textsc{\textbf{rel}} & \textsc{\textbf{jus}} & \textsc{\textbf{use}} \\
    \toprule
        \textsc{evaluators} & $0.721$ & $0.711$ & $0.632$ & $0.672$ \\
    \bottomrule
    \end{tabular}}
    \caption{Inter-annotator agreement statistics for the human evaluation task, measured using Krippendorff's Alpha with interval distance.}
    \label{tab:human_eval_agreement-all6_models}
\end{table}

\begin{table*}[t]
  \centering
  \small
  \adjustbox{max width=\textwidth}{
    \begin{tabular}{l||c||c||c|c|c||c|c|c|c}
    \toprule
    {} & \multicolumn{1}{c||}{\textsc{Length}} & \multicolumn{1}{c||}{\textsc{Abstractiveness}}  & \multicolumn{3}{c||}{\textsc{Auto Eval}} & \multicolumn{4}{c}{\textsc{Human Eval}}\\
    {} & \multicolumn{1}{c||}{\textsc{\# Tokens}} & \multicolumn{1}{c||}{\textsc{\%Novel Bigrams}} & \multicolumn{1}{c|}{\textsc{BLEU-4}} & \multicolumn{1}{c|}{\textsc{ROUGE-L}} & \textsc{BERTSc} & \textsc{FAC} & \textsc{REL} & \textsc{JUS} & \textsc{USE} \\
    \midrule
    \textsc{Annotators} & $28.9$ & $86.7\%$ & \multicolumn{3}{c||}{------} & $0.68$  & \bftab 4.43  & \bftab 0.92  & $0.77$ \\
    \midrule
    \textsc{ChatGPT} & $58.0$ & $81.8\%$ & \bftab 0.044 & $0.224$ & \bftab 0.347 & \bftab 0.88  & \bftab 4.42  & $0.85$  & \bftab 0.88 \\
    \textsc{Flan-T5} & $45.3$ & $16.0\%$ & $0.008$ & $0.066$ & $0.053$ & $0.44$ & $2.27$  & $0.25$  & $0.19$ \\
    \textsc{Alpaca-7b} & $48.6$ & $71.9\%$ & $0.040$ & \bftab 0.230 & $0.297$ & $0.57$  & $4.23$  & $0.79$  & $0.64$ \\
    \textsc{Alpaca-13b} & $19.7$ & $10.9\%$ & $0.003$ & $0.017$ & $0.066$ & $0.03$  & $1.13$  & $0.02$  & $0.02$ \\
    \textsc{Dolly-7b} & $79.7$ & $51.3\%$ & $0.020$ & $0.142$ & $0.157$ & $0.32$  & $2.44$  & $0.21$  & $0.18$ \\
    \textsc{Dolly-12b} & $73.3$ & $55.1\%$ & $0.024$ & $0.165$ & $0.199$ & $0.38$  & $2.79$  & $0.56$  & $0.38$ \\
    \bottomrule
    \end{tabular}}
  \caption{Experiment results from LLMs. We report the average performance across five independent runs. A more comprehensive report of the automatic metrics \textsc{BLEU-4}, \textsc{ROUGE-L}, and \textsc{BERTScore} is provided in Table \ref{tab:llm_results_full}, Appendix \secref{appendix:llm_results}.}
  \label{tab:llm_results-rationale-appendix-all6_models}
\end{table*}

The cognitive appraisal theories provide insights into the nature of the appraisal dimensions in distinguishing various emotions \cite{hofmann-etal-2020-appraisal, YeoOng2023Appraisals}: while different individuals may appraise the same situation distinctively, they are more likely to experience the same emotion when a consistent appraisal pattern emerges. For example, the cognitive dimension \textit{pls} (pleasantness) is often linked to joy, but unlikely to be associated with disgust \cite{smith1985patterns}. Therefore, specific emotions are hypothesized to stem from corresponding appraisal patterns \cite{YeoOng2023Appraisals}. By understanding how individuals appraise the situations they experience, we can subsequently make predictions regarding their emotional state. As a result, appraisal dimensions are valuable in differentiating emotional states, especially in cases where the emotions are highly interchangeable (e.g., \textit{disgust} and \textit{anger}).

Here, using the cognitive appraisal dimensions annotated in \dataset{}, we further explore and validate whether these appraisal dimensions alone are indicative of perceived emotions already annotated in {\sc CovidET}. While in the ideal scenario, both the appraisal and the objective event need to be present for emotion prediction, this small experiment will allow us to gauge which dimensions are more likely discriminative for a particular emotion. For each of the $7$ emotion classes labeled in \CovidET{}, we train a logistic regression model using the scales of the annotated $21$ applicable appraisal dimensions as features. We split \dataset{} using a random $80$:$20$ train-test partitioning, and aggregate the Likert-scale ratings for the $40$ posts that are doubly annotated by our human annotators following the aggregation setup discussed in \secref{sec:dataset}. We down-sample the training data for each logistic regression model to handle class imbalance issues. In addition, we encode the \textit{``not mentioned''} (NA) labels as an independent real-valued feature, and substitute their values with $0$. To prevent features of different scales or magnitudes from having a disproportionate influence on the models, we $Z$-normalize the scale ratings within each dimension for each annotator.

The F1 scores for each emotion using the trained logistic regression models on the test set are reported in Table \ref{tab:lr_f1}. We observe that the models are most capable at predicting emotions such as \textit{fear} and \textit{sadness}, whilst performing poorly on emotions \textit{disgust} and \textit{trust}. This is possibly due to the domain of our dataset: in \CovidET{}, \textit{fear} and \textit{sadness} are the most commonly found emotions whereas \textit{disgust} and \textit{trust} are scarcely present. On average, the classifiers achieve an average F1 of $0.22$ on the test set across all emotions.

To reveal the appraisal dimensions that are indicative of each emotion, we examine the weights from the trained logistic regression models. Specifically, we aim to validate the emotion appraisal dimensions that \citet{YeoOng2023Appraisals} identified to be predictive of emotions (including \textit{anger}, \textit{fear}, \textit{joy}, \textit{sadness}, and \textit{disgust}) from prior studies in psychology. In Table \ref {tab:desmond_emotion_prediction}, we show the appraisal dimensions found to be either positively predictive ($+$) or negatively predictive ($-$) of emotions. Please note that these indications are extracted from a recent meta-analysis from \citet{YeoOng2023Appraisals} with significance ($p<0.05$). In Table \ref{tab:desmond_emotion_prediction}, we highlight the indicative appraisal dimensions captured by our logistic regression models that are in line with \citet{YeoOng2023Appraisals}'s findings. We observe a certain degree of overlap between \citet{YeoOng2023Appraisals}'s identified emotion appraisal dimensions that are predictive of emotions and those captured by our logistic regression models. It should be noted that some appraisal dimensions may not be useful for all emotions included in Table \ref{tab:desmond_emotion_prediction}, since in \CovidET{} there are no Reddit posts annotated with neutral emotions: for example, as shown in Table \ref{tab:desmond_emotion_prediction}, \textit{crsp} (circumstances-responsibility) is found to be positively indicative for \textit{fear} and \textit{sadness}, while neutral for all other emotions. However, when compared to neutral emotions (i.e., in texts where no emotions are present), \textit{crsp} (circumstances-responsibility) may be a negative indicator for \textit{disgust}. Therefore, experimenting with \dataset{} may not reveal the extensive range of appraisal dimensions indicative of each emotion. Further investigations are needed to explore the predictability of these appraisal dimensions for emotions compared against neutral emotions.

\subsection{Topic Variations in Rationales} \label{appendix:lda}
We use Latent Dirichlet Allocation (LDA) \cite{10.5555/944919.944937} to extract topics from the natural language rationales annotated in \dataset{}. Stop-words such as common English function words and words that occur frequently in our instructions (e.g., \textit{narrator}, \textit{situation}) are removed prior to the topic modeling. The most prominent topic extracted by the LDA model for each dimension is shown in Table \ref{tab:rationales_lda}. We notice clear patterns of topics related to the appraisal dimension being assessed. For example, in dimension \textit{crsp} (circumstances-responsibility) we observe narrators of Reddit posts worrying about and blaming Delta, a COVID-19 variant, for causing the status quo, whereas in dimension \textit{fml} (familiarity) we note people are generally unfamiliar with the situation, as they are prone to seek advice and probe for information on the forum.

\subsection{An Example of Semantic Similarity} \label{appendix: semantic-similarity-example}
As discussed in \secref{sec:analysis}, commonly used automatic measures such as BLEU~\cite{papineni-etal-2002-bleu}, ROUGE~\cite{lin-2004-rouge}, and BERTScore~\cite{DBLP:journals/corr/abs-1904-09675} do not adequately capture semantic similarity in \dataset{}. Taking the post in Figure \ref{fig:dataset-example} for example. Both rationales for dimension $24$, namely \textit{``The narrator mentions how people who are vaccinated and mildly sick are still experiencing long COVID symptoms. They seem surprised by the continued COVID symptoms people are experiencing and how the situation seems to evolve.''} and \textit{``The narrator really didn't expect this situation since they mention being able to taste freedom, believing the pandemic is ending, when suddenly they heard news that vaccinated people are still getting long covid and now they think the pandemic will never end.''} convey the reasons for why the narrator fails to expect the situation to occur. However, the automatic metrics reveal low agreement between these two rationales, with a \textsc{BLEU-4} score of $0.018$, \textsc{ROUGE-L} of $0.231$, and a re-scaled \textsc{BERTScore} of $0.237$. This finding is in line with work showing the challenges of \emph{evaluating} generation~\cite{gehrmann2021gem,celikyilmaz2020evaluation}; we similarly conclude that automatic evaluation metrics may poorly reflect the correctness of a rationale for a subjective emotion appraisal dimension.

\section{Prompt Templates}\label{appendix:prompt-templates}
\begin{figure*}
    \centering
    \includegraphics[width=\textwidth]{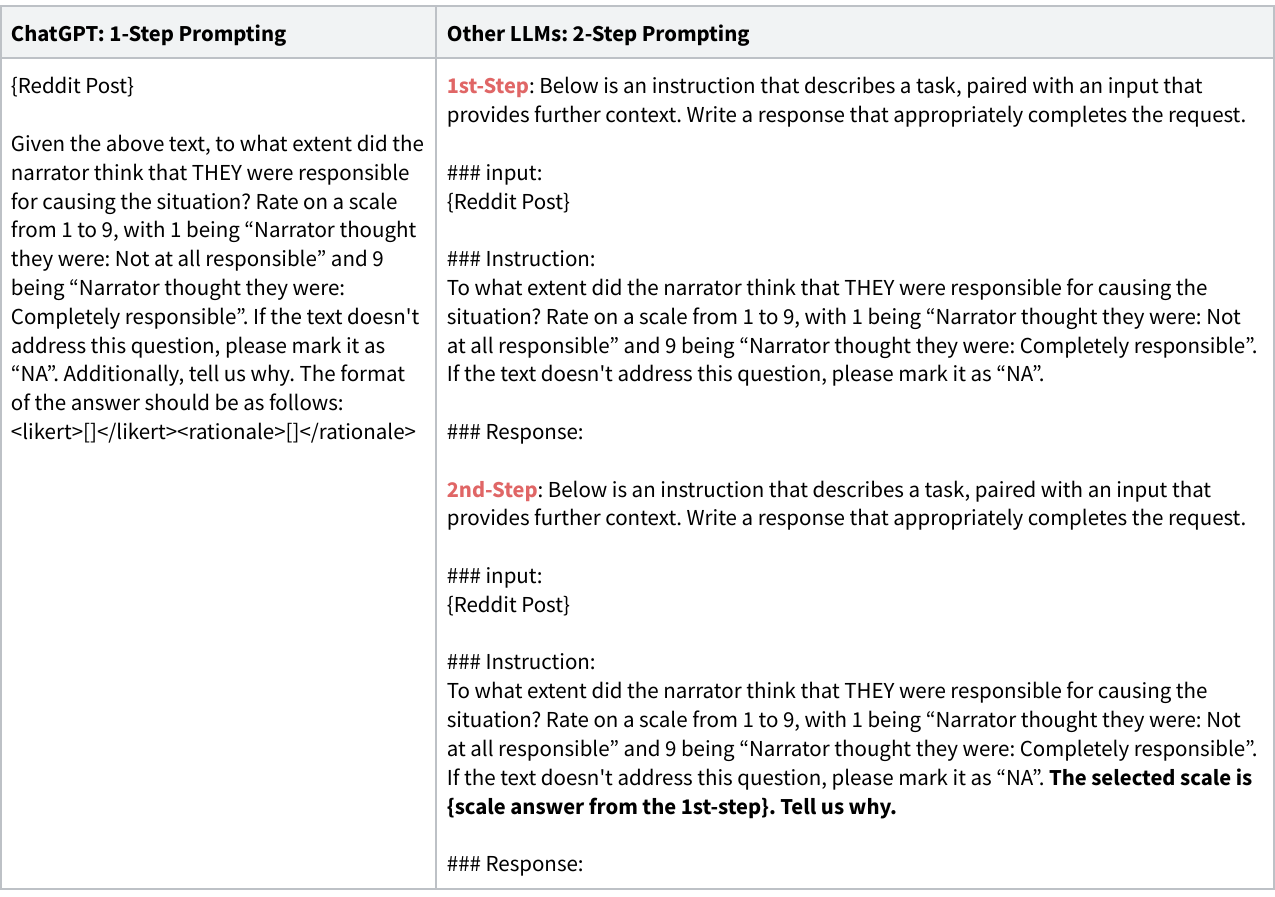}
    \caption{Prompt templates (taking dimension 1 as an example).}
    \label{fig:prompt_templates}
\end{figure*}

The templates for prompting the LLMs are shown in Figure \ref{fig:prompt_templates}. We use ``1-step'' prompting to elicit both a rating and a rationale with a single prompt from ChatGPT. For all other language models, we apply ``2-step'' prompting, which first elicits the rating for the appraisal dimension, then conditioned on the response for the rating we further elicit the rationale for the selection.

\section{Full LLM Rationale Measures}\label{appendix:llm_results}

\paragraph{Rationale Automatic Evaluation.}

We provide the full statistics of the automatic rationale agreement measured using BLEU \cite{papineni-etal-2002-bleu}, ROUGE \cite{lin-2004-rouge}, and BERTScore \cite{DBLP:journals/corr/abs-1904-09675} for the \textbf{\textit{all $6$} LLMs' responses} against the gold annotations in Table \ref{tab:llm_results_full}.

As discussed in \secref{subsec:auto_eval}, ChatGPT is the most performant language model in providing natural language rationales, with values from these metrics comparable to those between different rationales from our two annotators. Alpaca-7B also achieves comparable performance in these automatic measures, despite its relatively poor capability in terms of selecting Likert-scale ratings.

In addition, we observe that other language models such as FLAN-T5 and Dolly lag behind considerably compared to ChatGPT and Alpaca-7B. Enchantingly, the automatic metrics suggest that Alpaca-13B is the worst language model among our LLMs under assessment, with a markable degradation from Alpaca-7B. Further investigation reveals that Alpaca-13B tends to respond with ``\textit{Tell us why.</s>}'' when prompted to generate the natural language rationale for the Likert-scale rating it selects, which takes up more than $84\%$ of its rationale responses. The debasement of the Alpaca model in spite of the increase in the model's scale raises questions regarding the scaling law in our current task of appraising cognitive emotion dimensions in context.

\paragraph{Rationale Human Evaluation.}

\begin{figure}[t]
    \centering
    \includegraphics[width=\columnwidth]{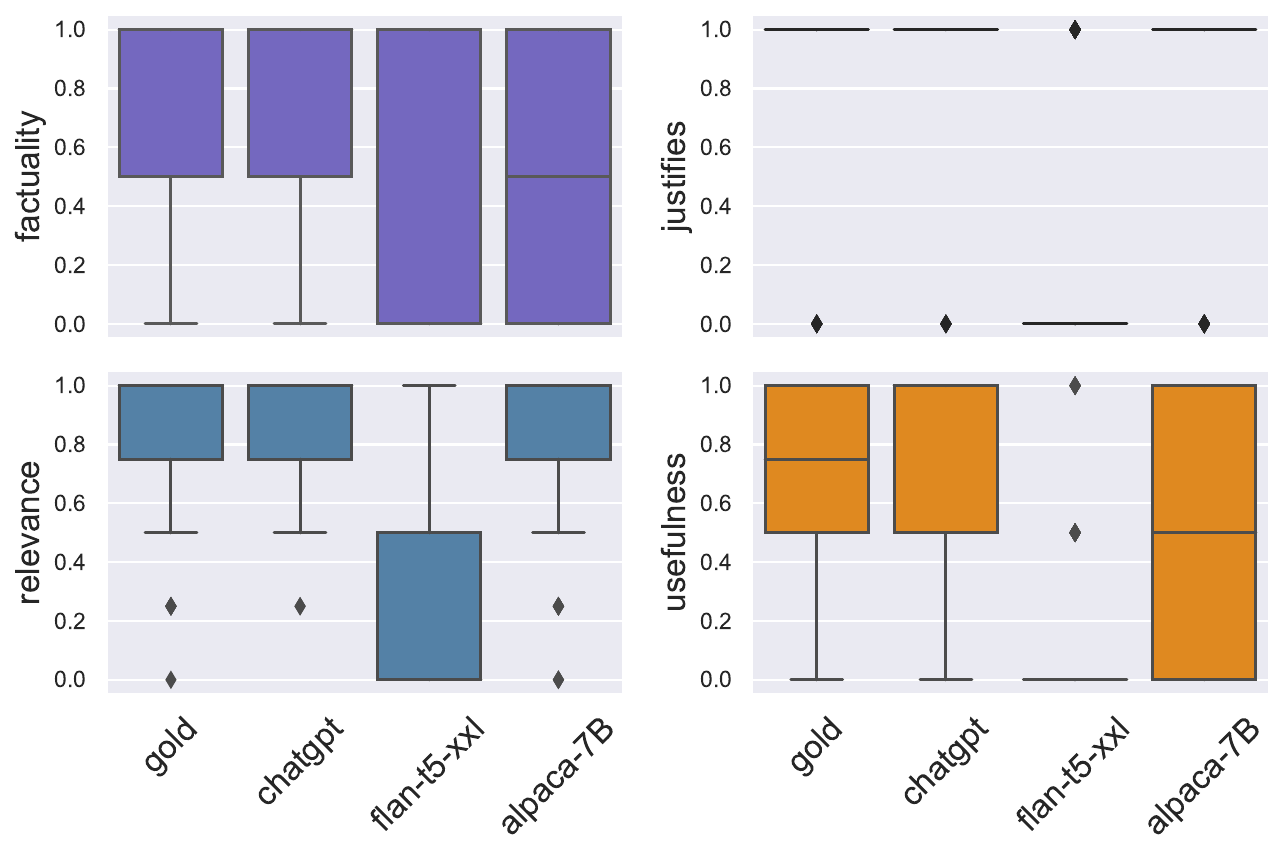}
    \caption{Box plots of the results from the human evaluation task for \textit{the most-performant $3$} LLMs (i.e., ChatGPT, Alpaca-7B, and FLAN-T5).}
    \label{fig:human_eval_results-top_3}
\end{figure}

We provide the box plots of the results from the human evaluation for \textit{the most-performant $3$} language models (i.e., ChatGPT, Alpaca-7B, and FLAN-T5) in Figure \ref{fig:human_eval_results-top_3}.

Furthermore, we also provide the results for the human evaluation regarding \textbf{\textit{all $6$} LLMs} assessed in this paper. Following the setup in \secref{subsec:human_eval}, we evaluate and analyze LLM-generated rationales when the model made a near-correct prediction of the Likert-scale rating for that particular dimension compared against the gold human ratings. Specifically, we sample the \textit{intersection} of dimensions (post, dimension) tuples where \textit{all $6$} LLMs' (i.e., ChatGPT, FLAN-T5, Alpaca-7B, Alpaca-13B, Dolly-7B, and Dolly-12B) ratings fall in the range of an absolute difference of $1$ to \emph{one of} the annotated scale-ratings. This results in $30$ rationales annotated by human annotators and $26$ natural language rationales from each LLM. We report the inter-evaluator agreement using Krippendorff's Alpha with interval distance in Table \ref{tab:human_eval_agreement-all6_models}, which shows substantial agreement~\cite{artstein-poesio-2008-survey} across all criteria.

\begin{figure}[t]
    \centering
    \includegraphics[width=\columnwidth]{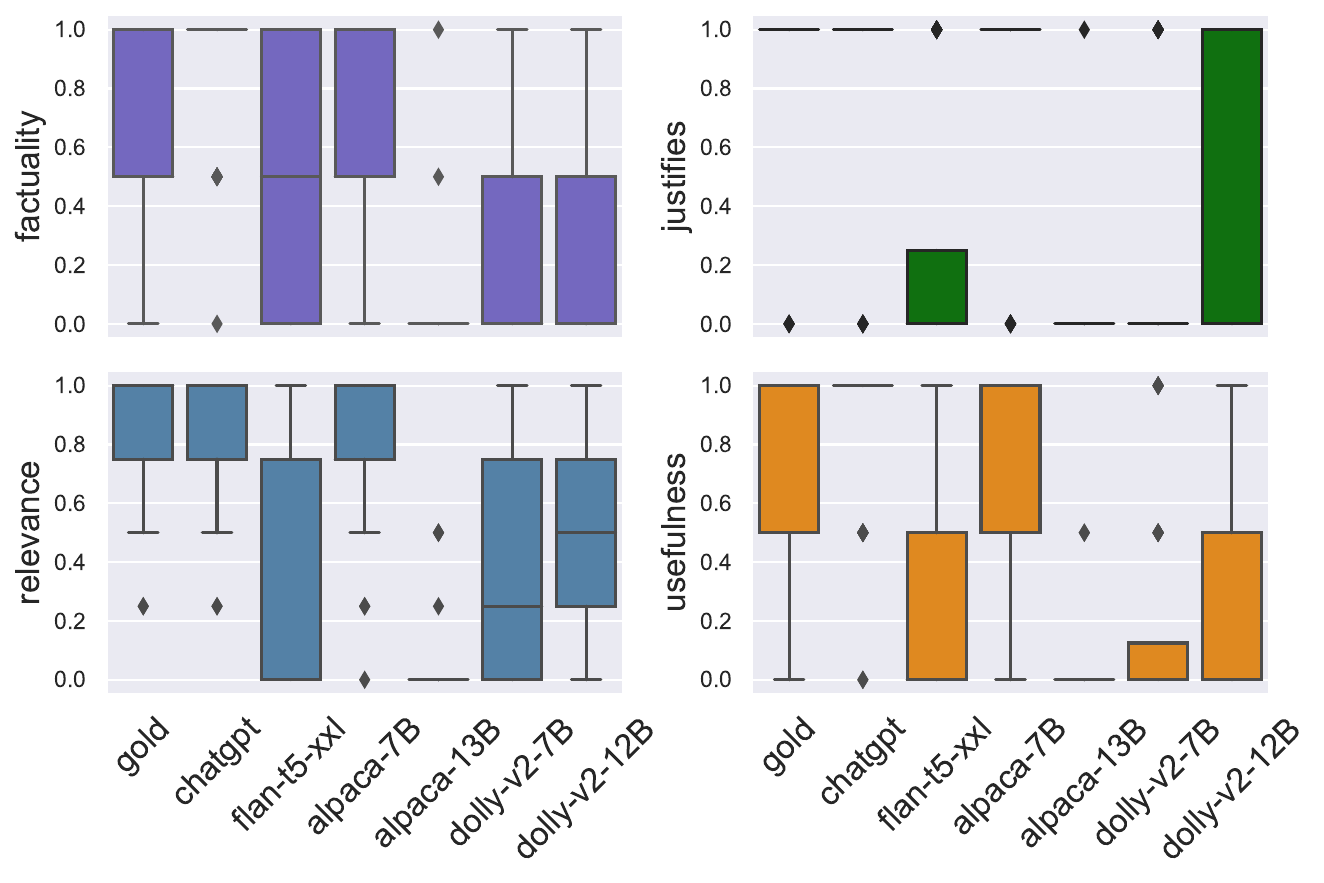}
    \caption{Box plots of the results from the human evaluation task for \textit{all $6$} LLMs.}
    \label{fig:human_eval_results-all_6}
\end{figure}

Results from the human evaluation for \textit{all $6$} LLMs are reported in Table \ref{tab:llm_results-rationale-appendix-all6_models}. We observe that apart from ChatGPT and Alpaca-7B, all other LLMs including FLAN-T5, Alpaca-13B, Dolly-7B, and Dolly-12B achieve similarly low performance on providing natural language rationales for cognitive emotion appraisals. We provide the box plots of the results from the human evaluation for \textit{all $6$} language models in Figure \ref{fig:human_eval_results-all_6}.

\begin{table}[t]
    \small
    \centering
    \adjustbox{max width = \columnwidth}{
    \begin{tabular}{l}
        \toprule
            \multicolumn{1}{c}{\textbf{\textsc{Reddit Post}}} \\
            \multicolumn{1}{p{\columnwidth}}{I want to clarify that I am in no way shape an anti masker, I believe we do need them during the pandemic and any future outbreaks. However, the masks are a nightmare for me personally. They trigger off my migraines, I could hardly concentrate at school because of it. I also have a sensory disability, which makes wearing masks 10 times harder. Everything is brighter, louder, and everything feels more crowded when I wear them. Its horrible. And it looks like we have to wear them for the next school year, and its my final year. I cant risk it. So I want to know if any you know where I can get masks that are migraine/sensory friendly? While still doing the job a normal medical mask does. Thanks.} \\
        \midrule
            \multicolumn{1}{c}{\textcolor{blue}{\textbf{\textsc{Appraisal Dimension \textit{ocrl}}}}} \\
            \multicolumn{1}{p{\columnwidth}}{\textcolor{blue}{To what extent did the narrator think that OTHER PEOPLE were controlling what was happening in the situation?}} \\
        \midrule
            \multicolumn{1}{c}{\textbf{\textsc{Human Annotations}}} \\
            \multicolumn{1}{p{\columnwidth}}{$\rightarrow$ \textsc{Scale}: 1} \\
            \multicolumn{1}{p{\columnwidth}}{$\rightarrow$ \textsc{Rationale}: \emph{The narrator does not mention anyone else as controlling what's happening in the situation.}} \\
            \multicolumn{1}{l}{\hspace{0.25\columnwidth}\textbf{\textcolor{red}{\textsc{Factuality}: $0.25$ \quad (Minor Error $+$ No)}}} \\
            \multicolumn{1}{l}{\hspace{0.25\columnwidth}\textbf{\textcolor{red}{\textsc{Usefulness}: $0.50$ \quad (Maybe $+$ Maybe)}}} \\
        \midrule
            \multicolumn{1}{c}{\textbf{\textsc{ChatGPT's Responses}}} \\
            \multicolumn{1}{p{\columnwidth}}{$\rightarrow$ \textsc{Scale}: 1} \\
            \multicolumn{1}{p{\columnwidth}}{$\rightarrow$ \textsc{Rationale}: \emph{The text does not suggest that other people are controlling the situation. The narrator is simply seeking advice on where to find masks that are more suitable for their needs.}} \\
            \multicolumn{1}{l}{\textbf{\textcolor{red}{\hspace{0.25\columnwidth}\textsc{Factuality}: $0.75$ \quad (Yes $+$ Minor Error)}}} \\
            \multicolumn{1}{l}{\textbf{\textcolor{red}{\hspace{0.25\columnwidth}\textsc{Usefulness}: $0.75$ \quad (Yes $+$ Maybe)}}} \\
        \bottomrule
    \end{tabular}}
    \caption{An example found in our human evaluation where ChatGPT scores higher than our human annotators on metrics \textit{factuality} and \textit{usefulness}.}
    \label{tab:human_chatgpt_rationale_ex}
\end{table}

\section{Model Responses Analyses}\label{appendix:model_analyses}
\begin{figure*}[htbp]
    \centering
    \begin{subfigure}[b]{\textwidth}
        \includegraphics[width=0.48\textwidth]{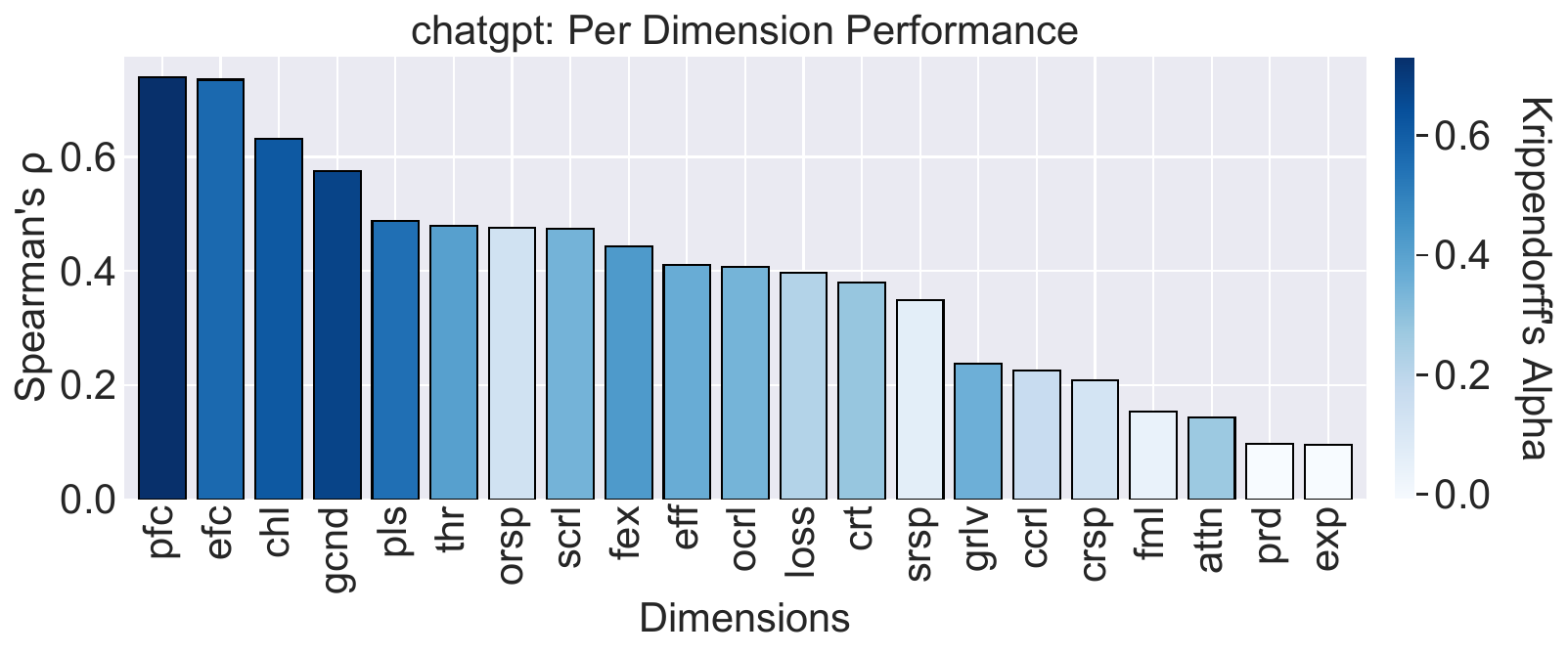}
        \hfill
        \includegraphics[width=0.48\textwidth]{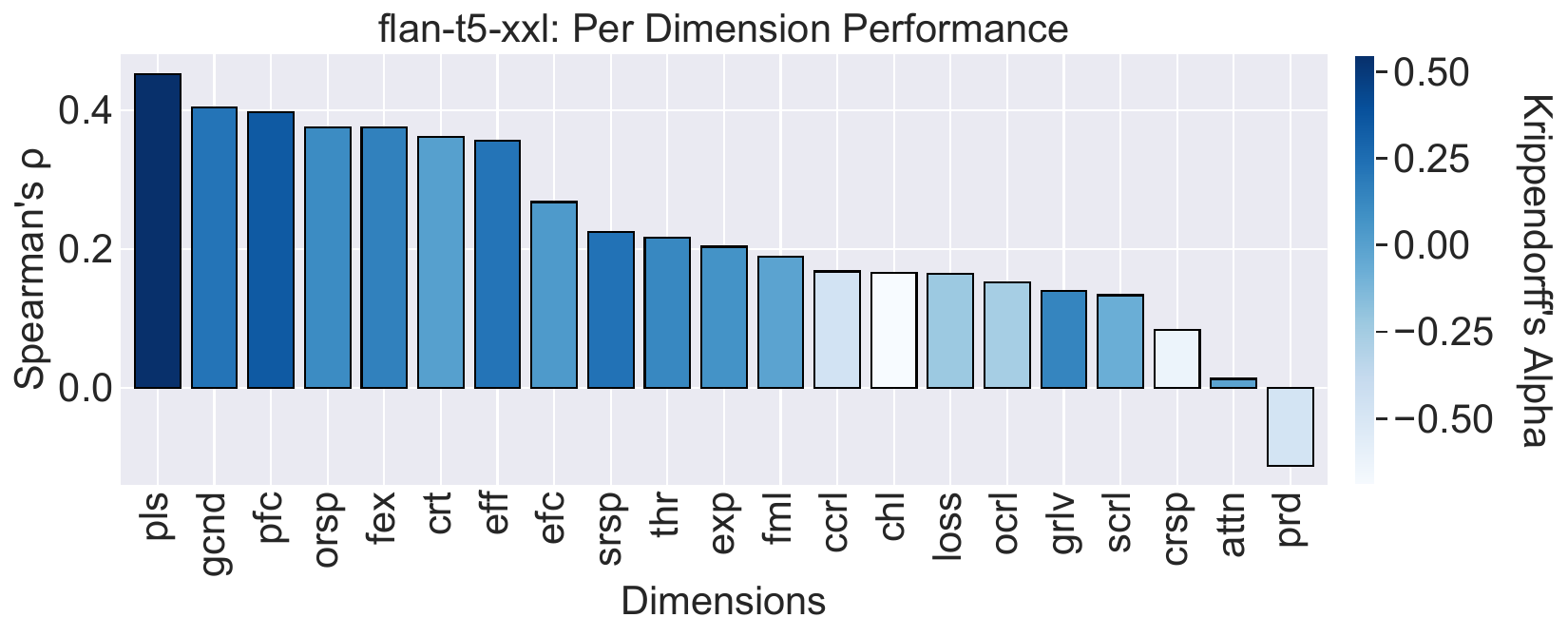}
    \end{subfigure}

    \medskip

    \begin{subfigure}[b]{\textwidth}
        \centering
        \includegraphics[width=0.48\textwidth]{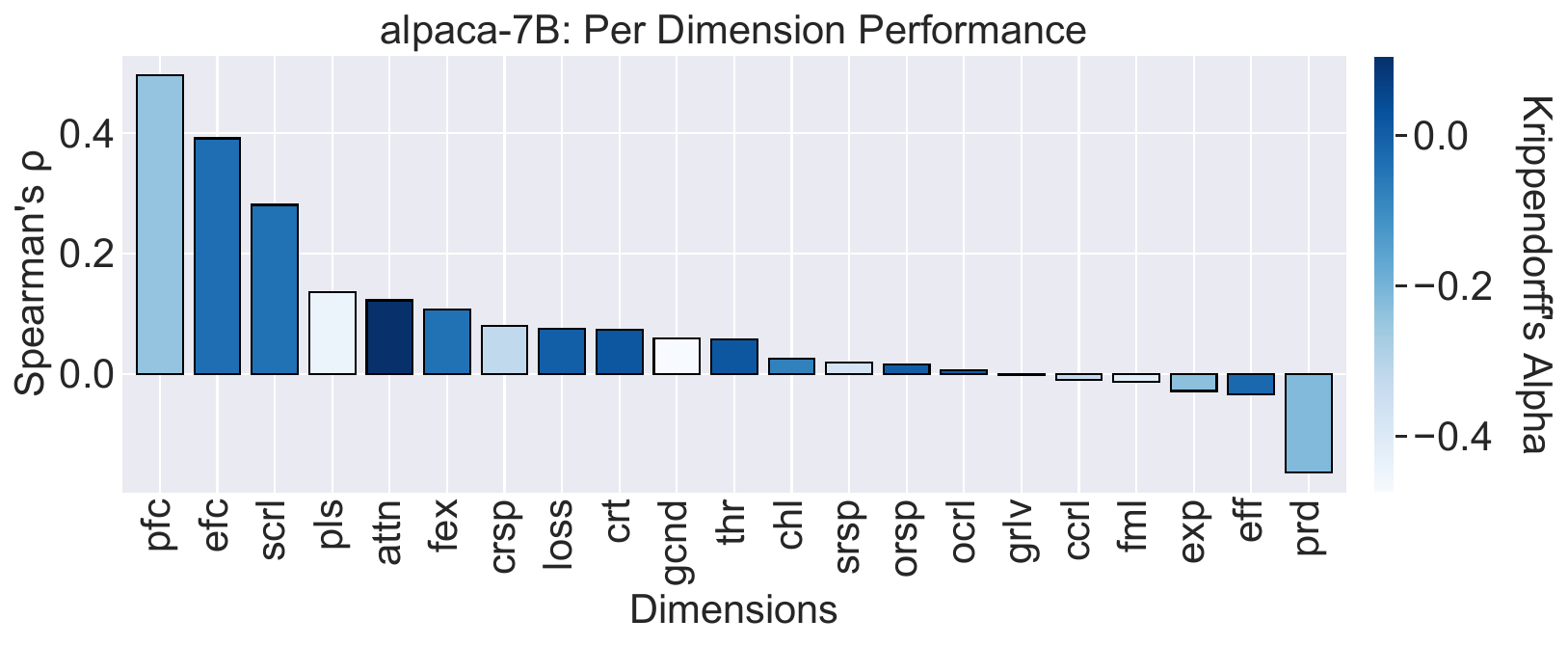}
        \hfill
        \includegraphics[width=0.48\textwidth]{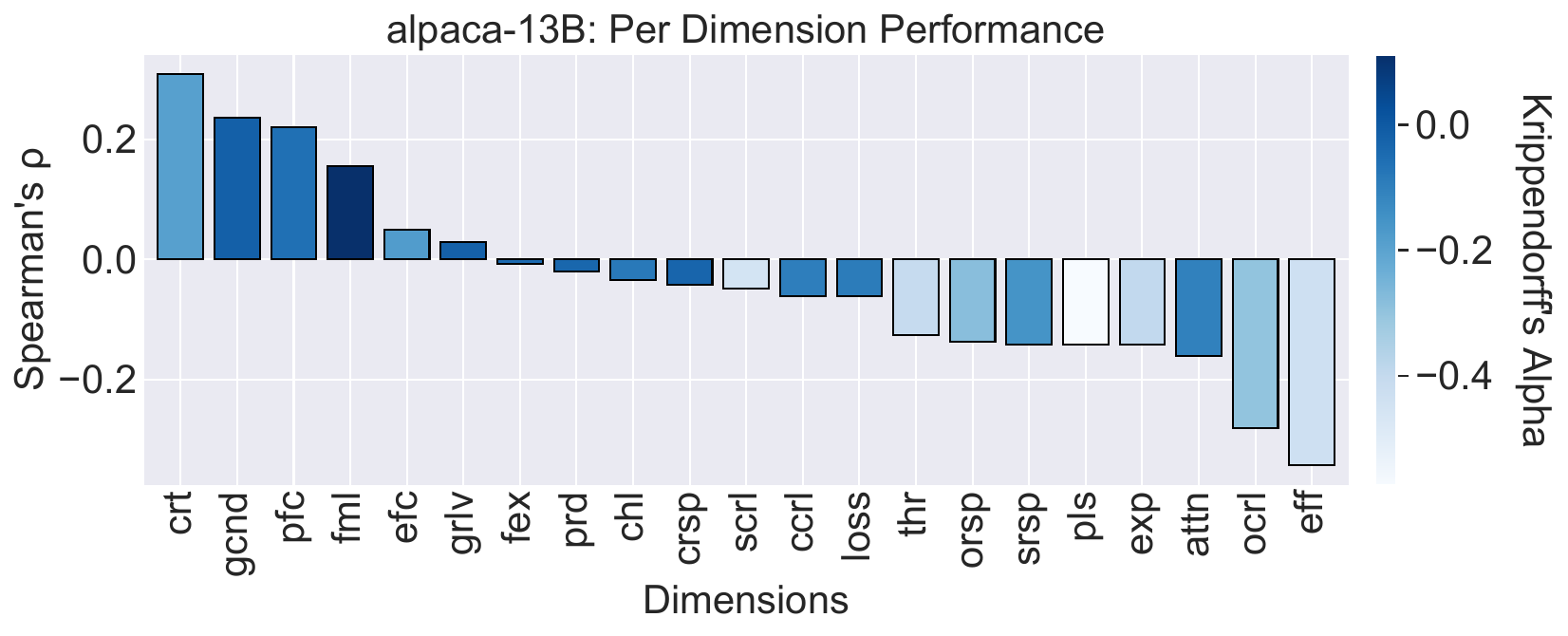}
    \end{subfigure}

    \medskip

    \begin{subfigure}[b]{\textwidth}
        \centering
        \includegraphics[width=0.48\textwidth]{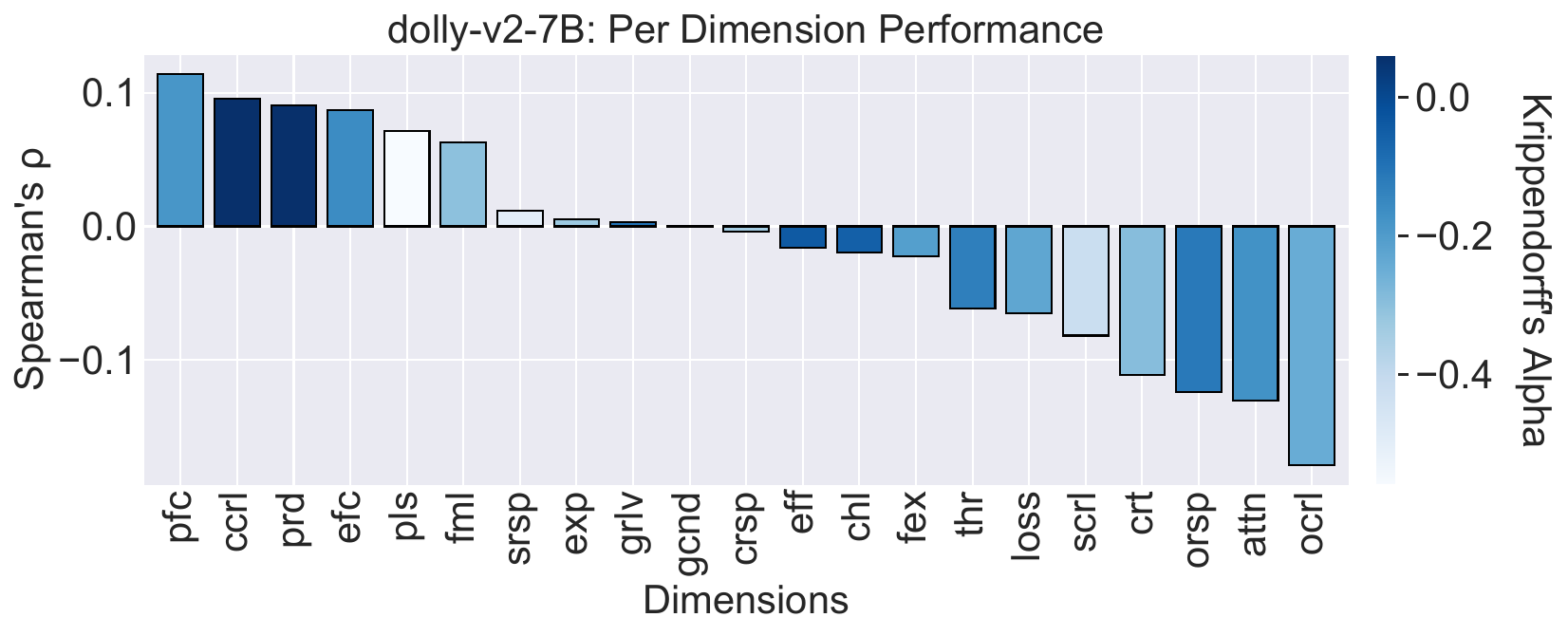}
        \hfill
        \includegraphics[width=0.48\textwidth]{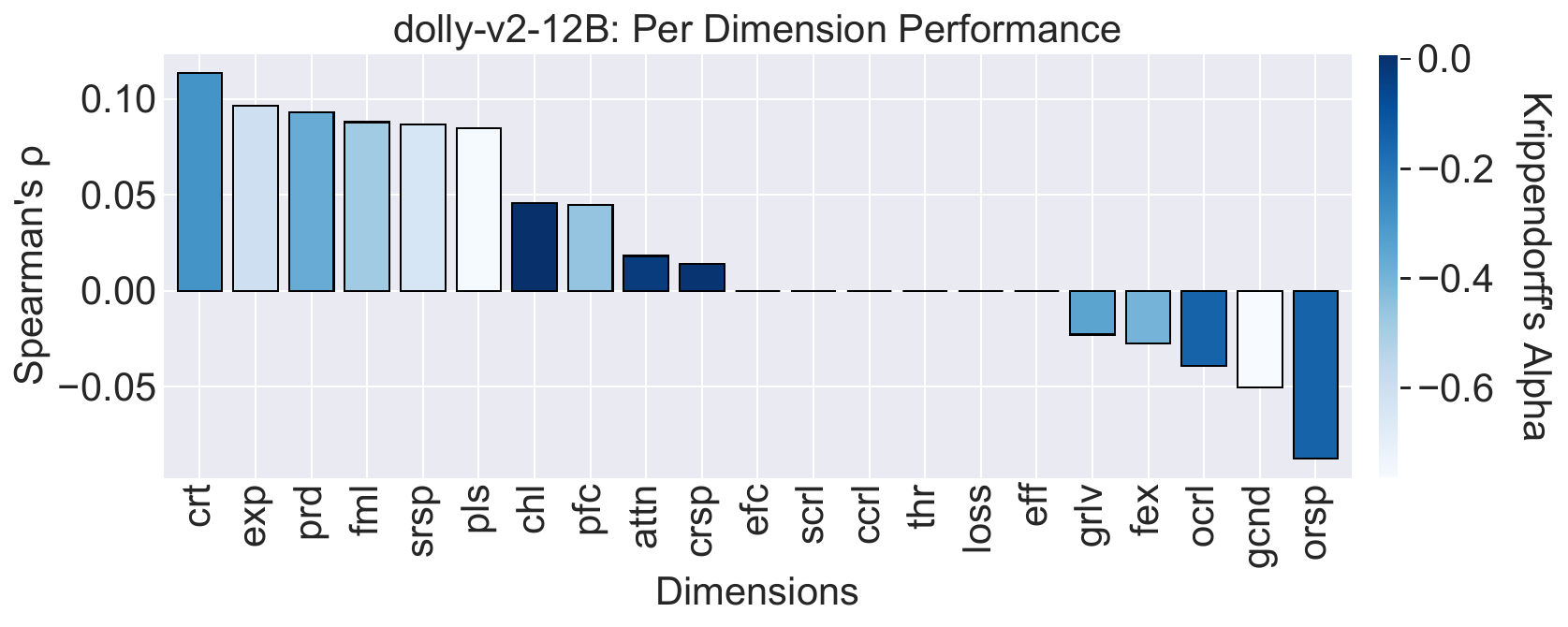}
    \end{subfigure}

    \caption{LLMs' performance in terms of Spearman correlation and Krippendorff's alpha (using interval distance) against the gold annotations within each group of dimensions (averaged performance across $5$ independent runs).}
    \label{fig:llm_pearson_per_group}
\end{figure*}

\begin{figure*}[htbp]
    \centering
    \begin{subfigure}[b]{\textwidth}
        \centering
        \includegraphics[width=0.48\textwidth]{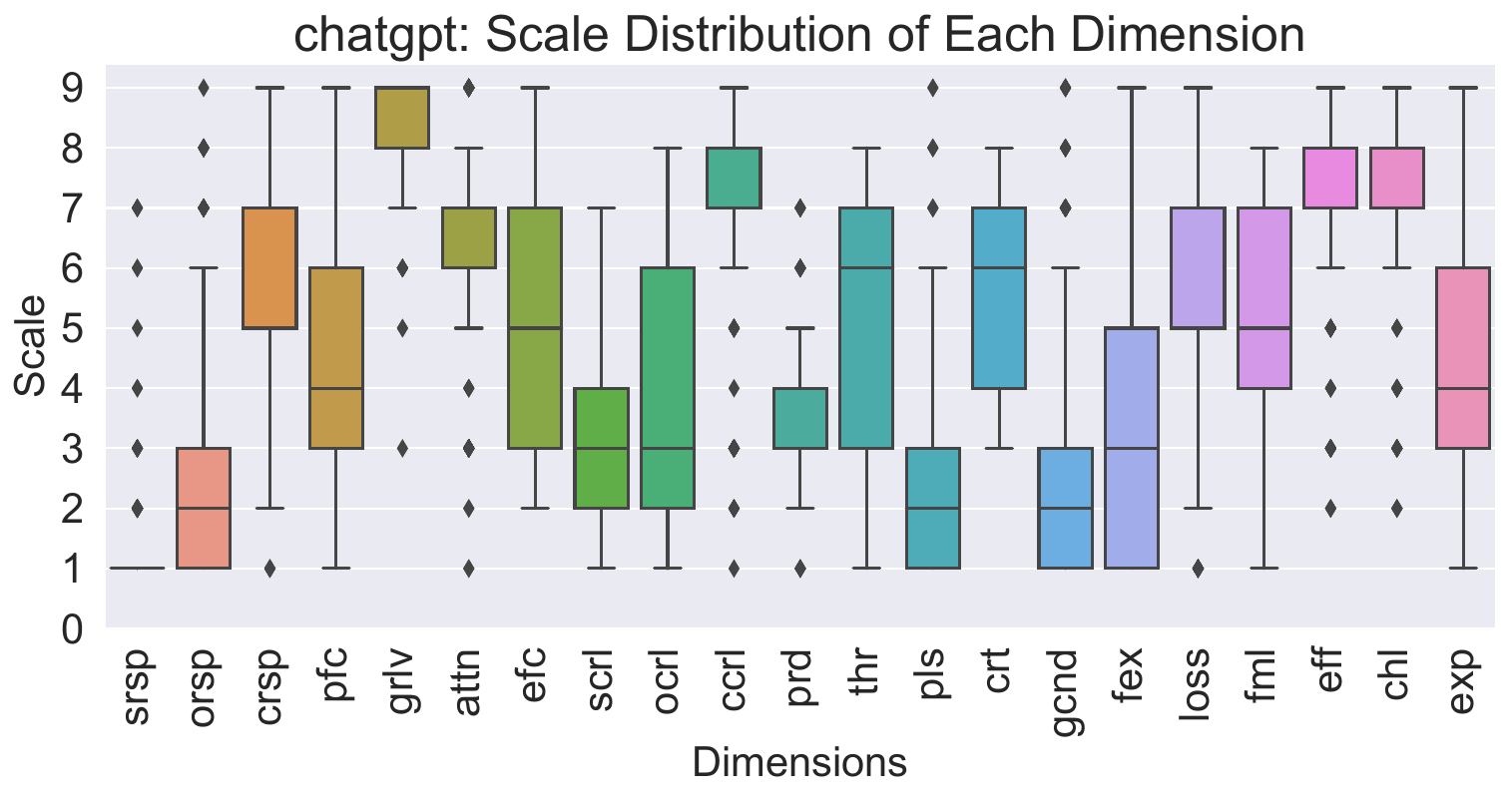}
        \hfill
        \includegraphics[width=0.48\textwidth]{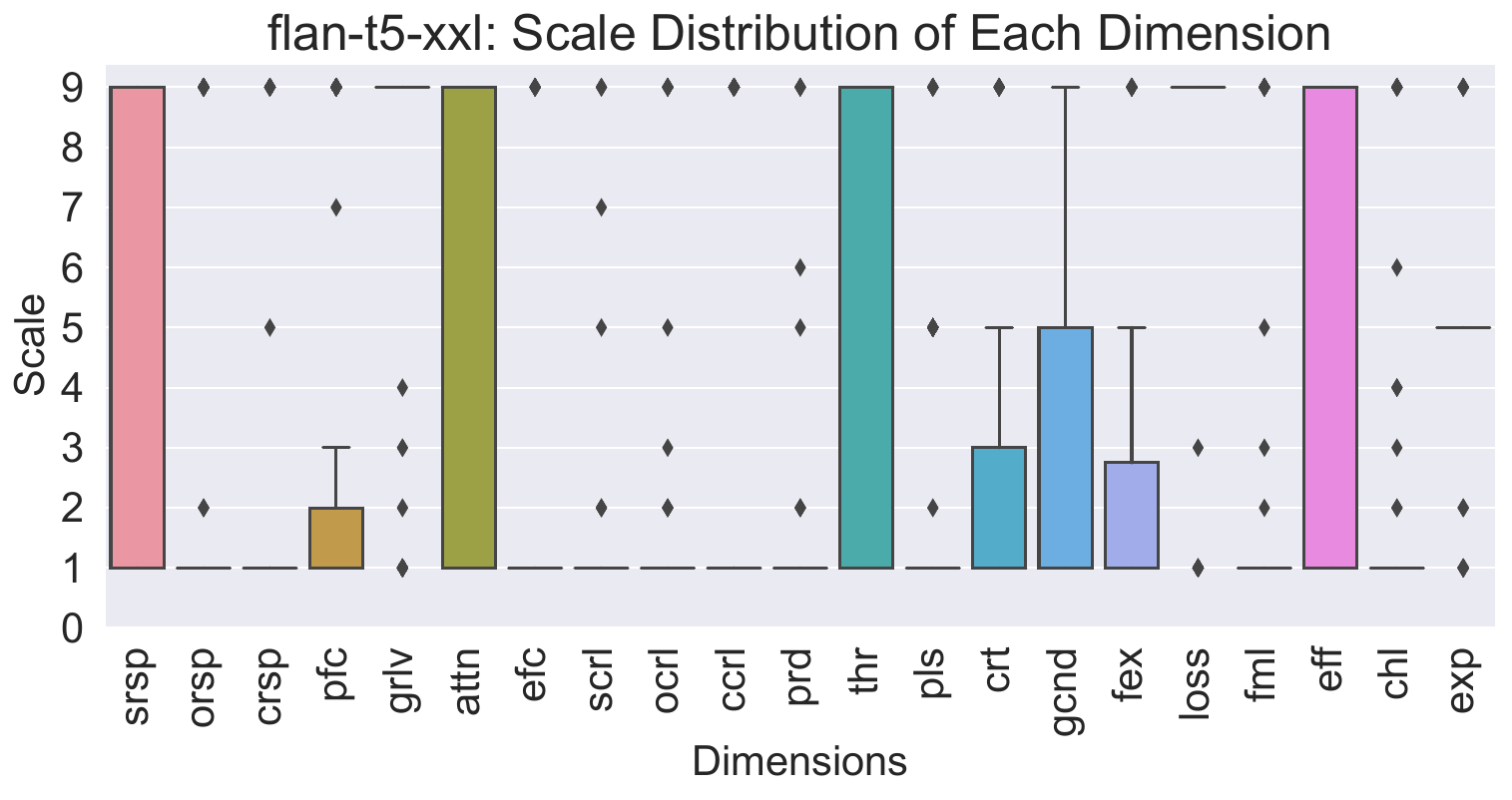}
    \end{subfigure}

    \medskip

    \begin{subfigure}[b]{\textwidth}
        \centering
        \includegraphics[width=0.48\textwidth]{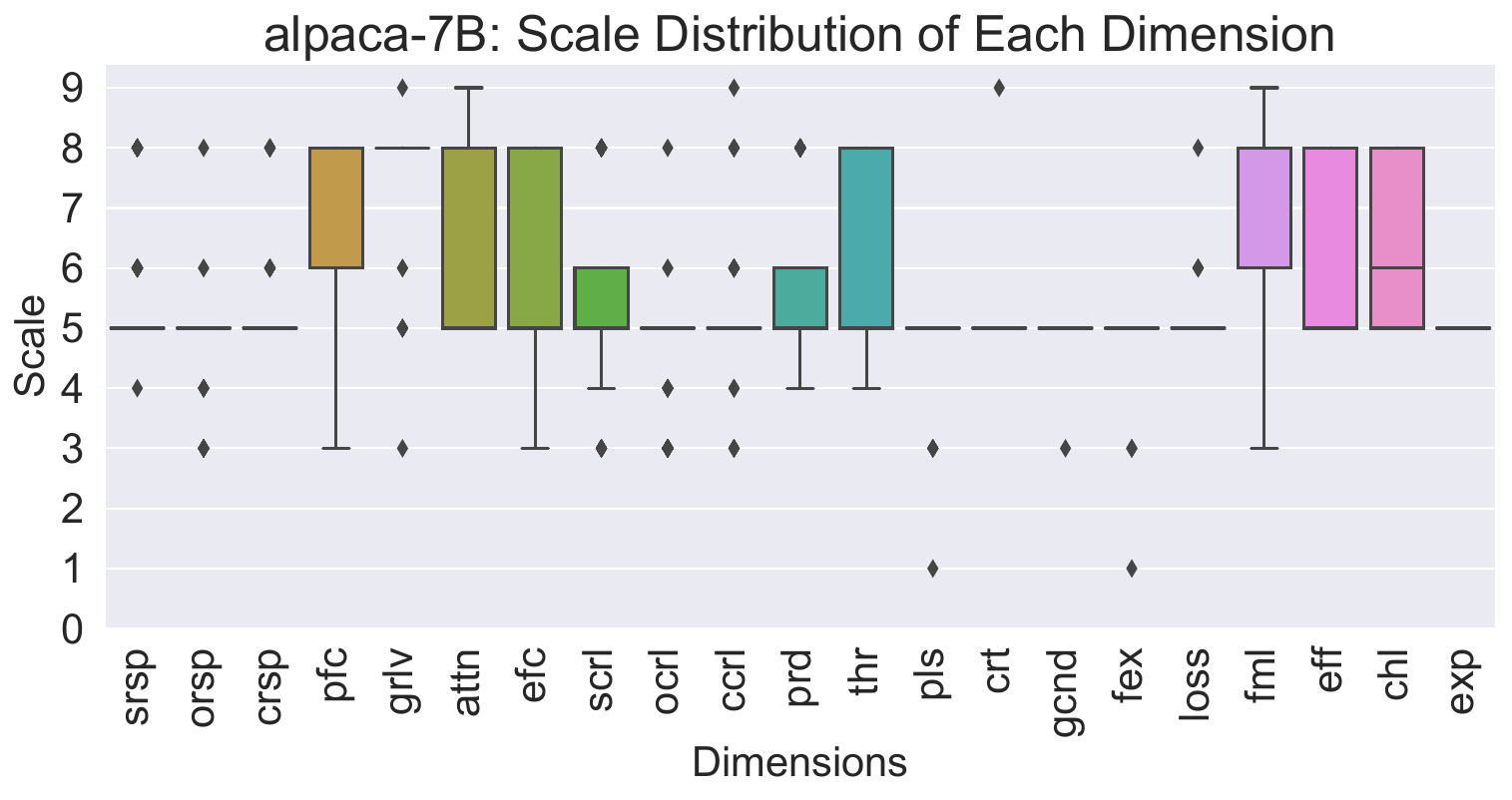}
        \hfill
        \includegraphics[width=0.48\textwidth]{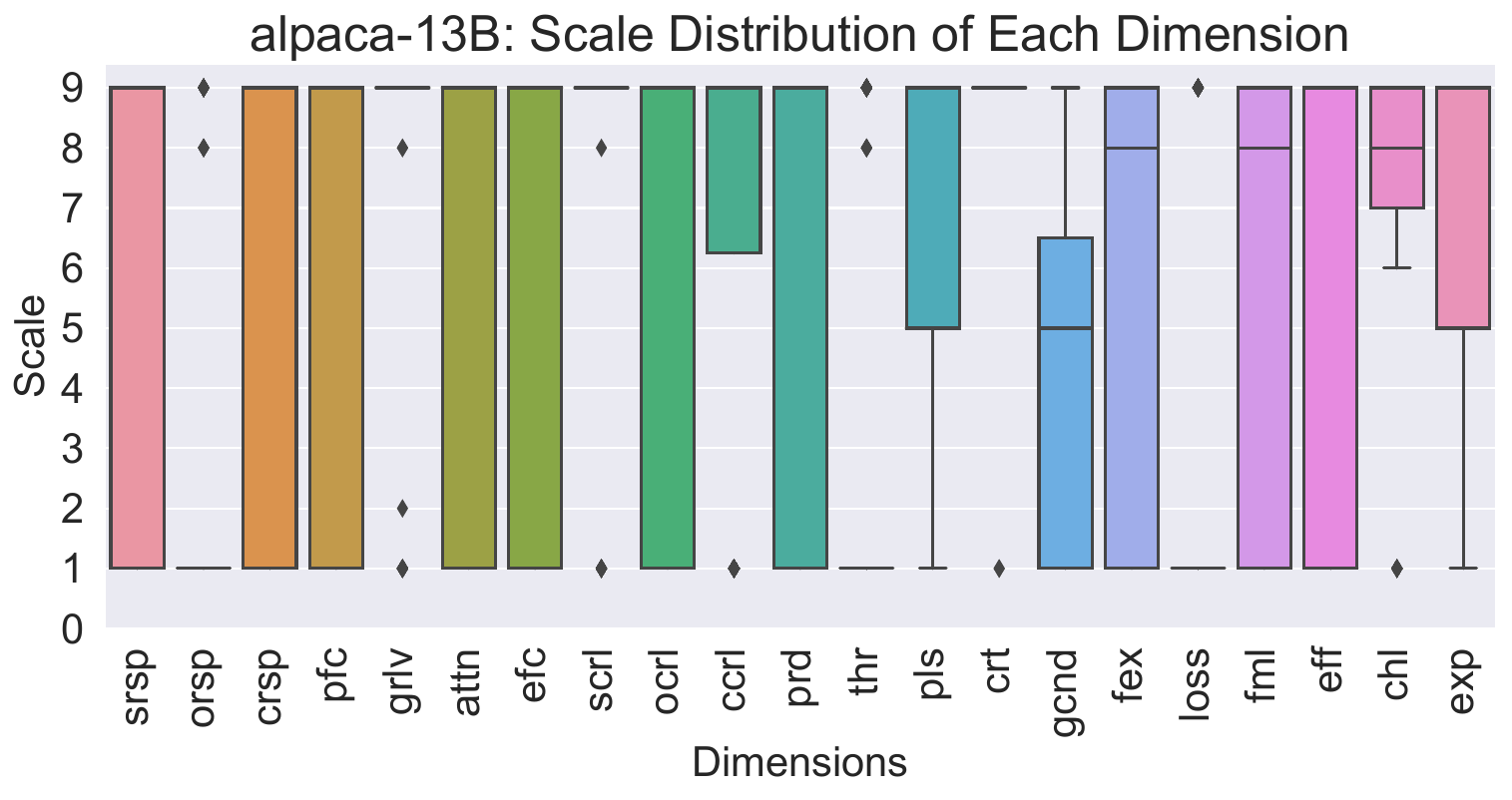}
    \end{subfigure}

    \medskip

    \begin{subfigure}[b]{\textwidth}
        \centering
        \includegraphics[width=0.48\textwidth]{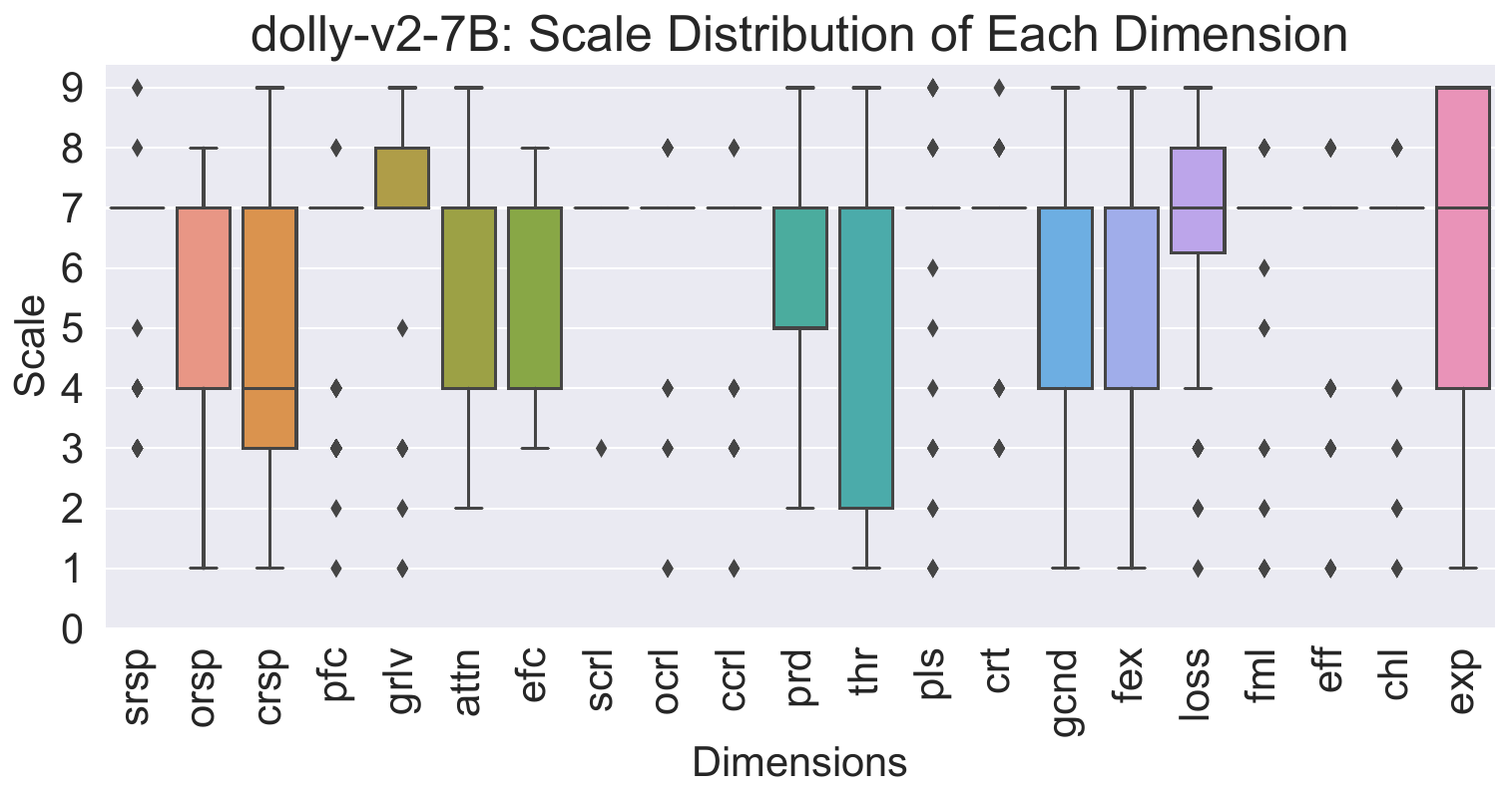}
        \hfill
        \includegraphics[width=0.48\textwidth]{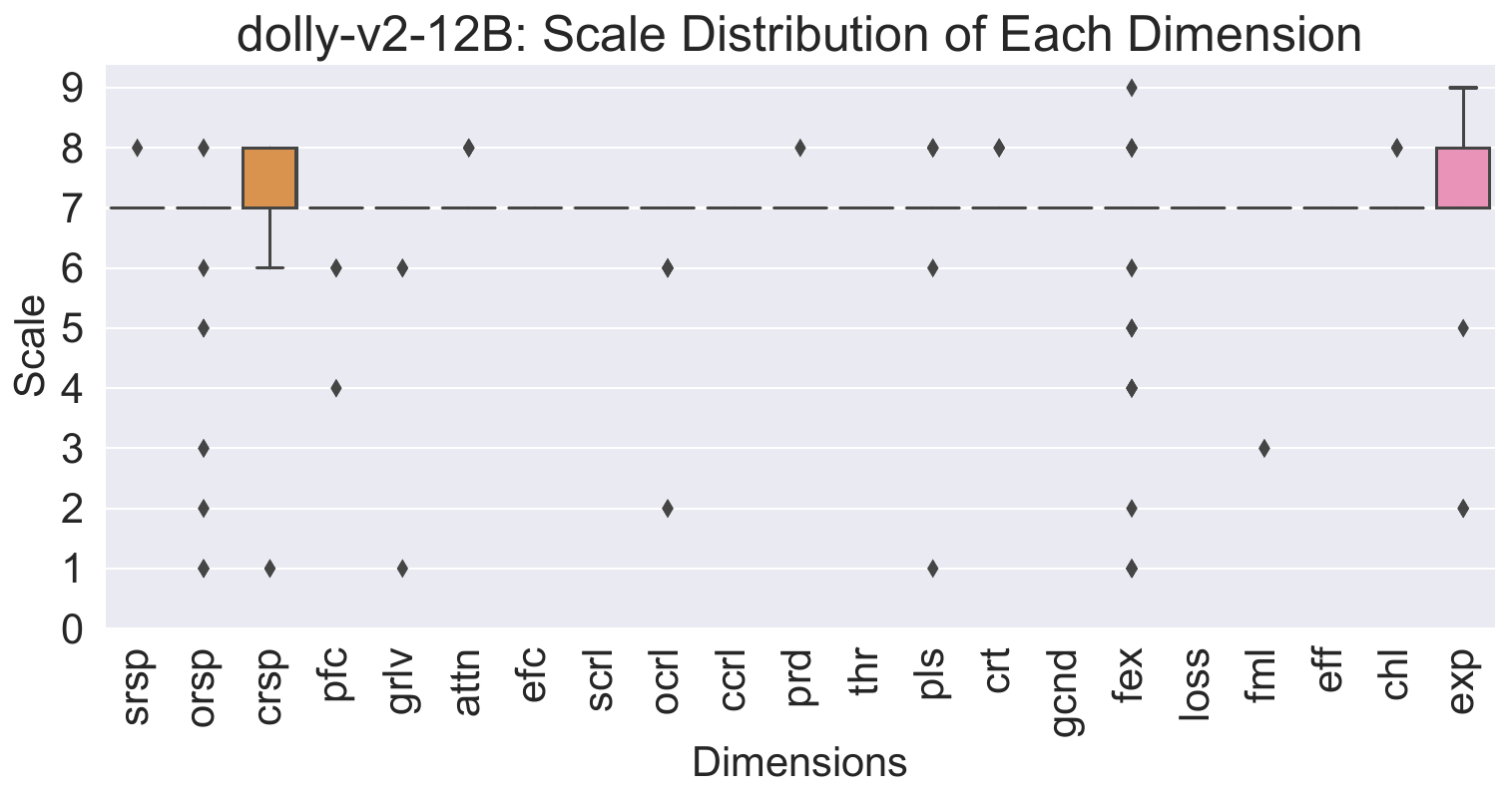}
    \end{subfigure}
    \caption{Box-plots for the LLMs' Likert-scale rating responses, measured across $5$ independent runs.}
    \label{fig:llm_boxplot}
\end{figure*}

The LLMs' performance in terms of Likert-scale rating selections measured using Spearman correlation and Krippendorff's alpha against the gold annotations are shown in Figure \ref{fig:llm_pearson_per_group}. Additionally, the box plots for each LLM's Likert-scale ratings are shown in Figure \ref{fig:llm_boxplot}.

\section{Human Evaluation Framework}\label{appendix:evaluation-framework}
\begin{figure*}[htpb]
  \centering
  \includegraphics[width=\textwidth, page=1]{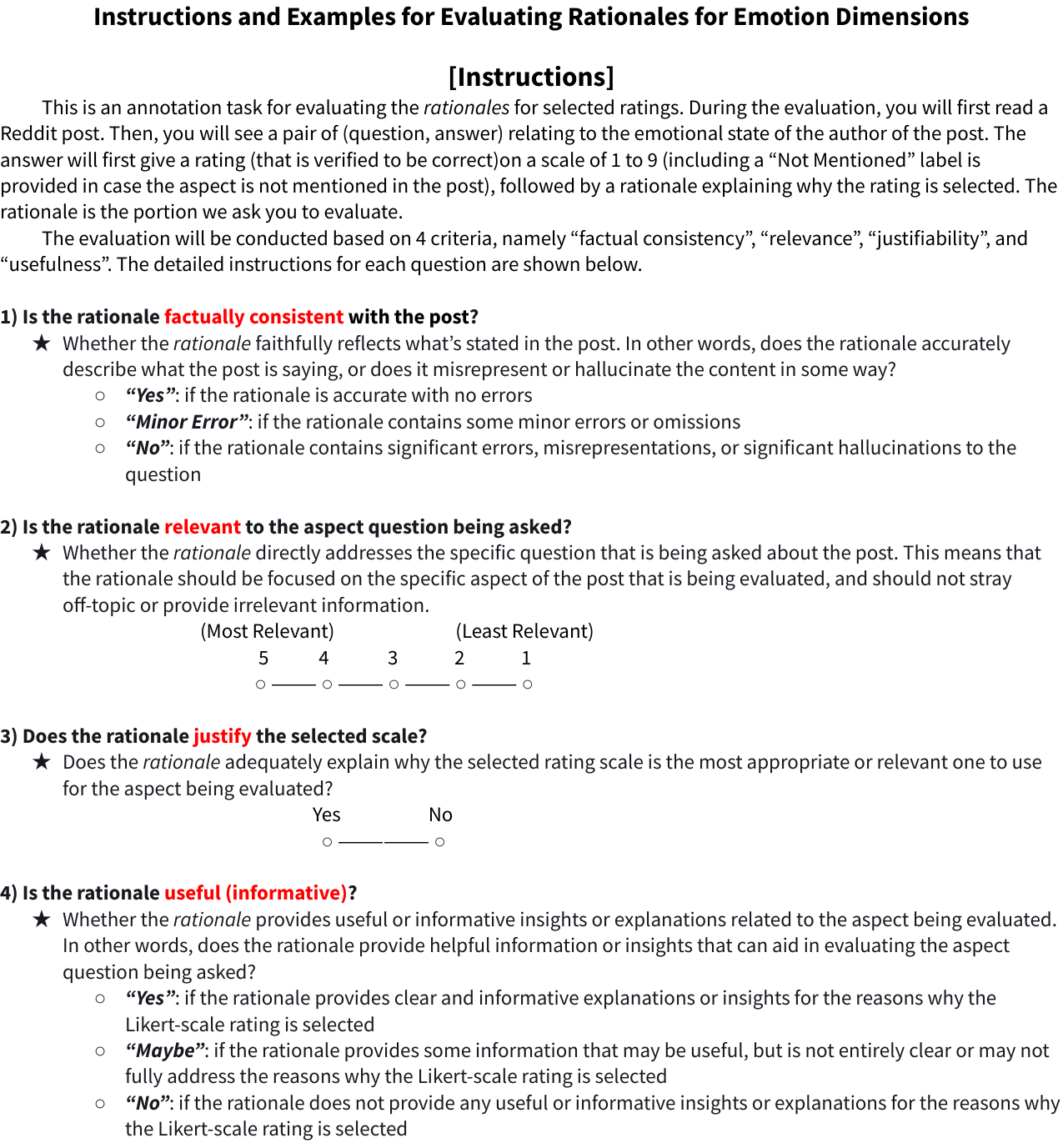}
  \caption{Instructions for the human evaluation described in \secref{subsec:human_eval} (Page $1/2$).}
  \label{fig:human_eval_instructions1}
\end{figure*}

\begin{figure*}[htpb]
  \centering
  \includegraphics[width=\textwidth, page=2]{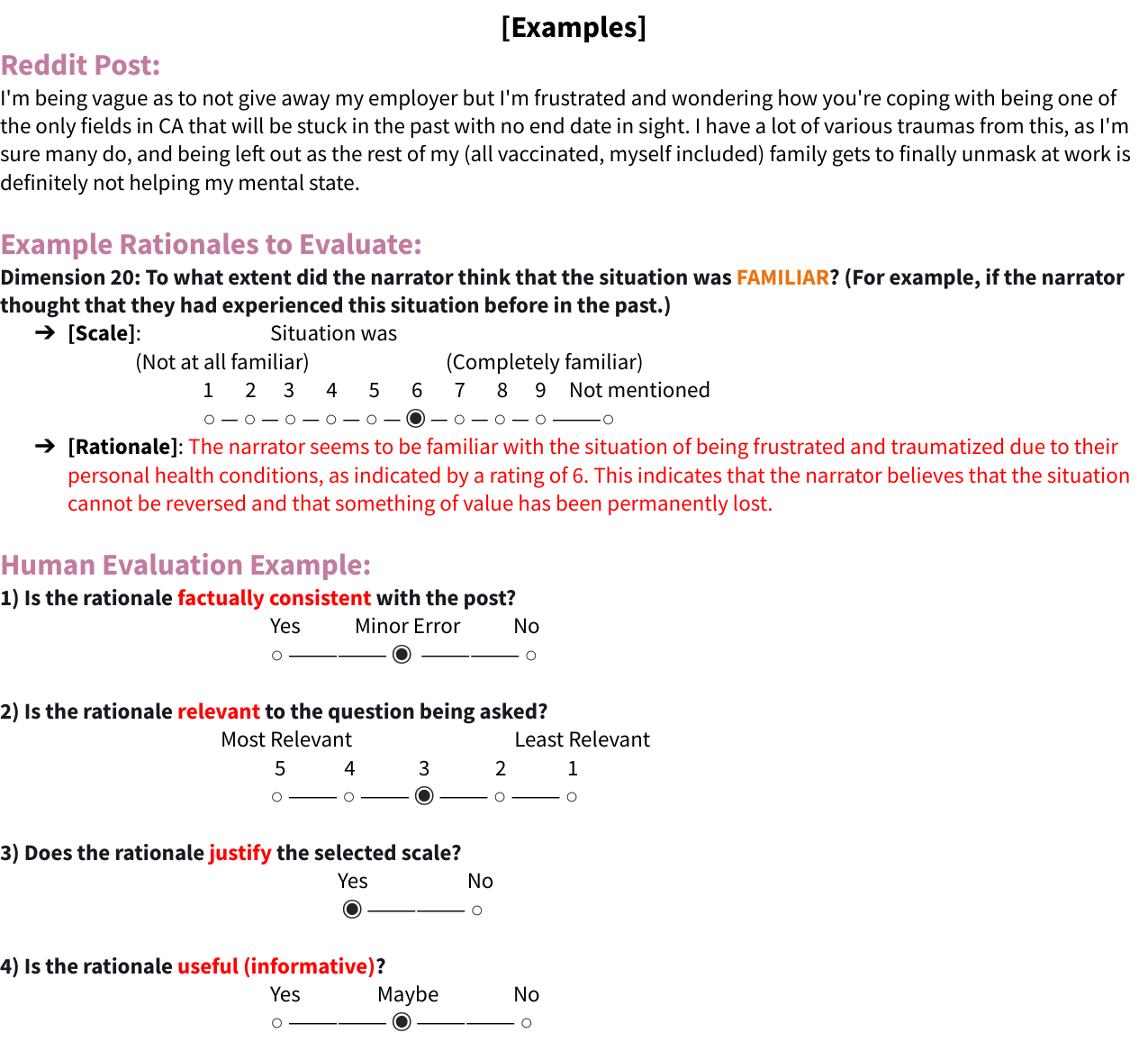}
  \caption{Instructions for the human evaluation described in \secref{subsec:human_eval} (Page $2/2$).}
  \label{fig:human_eval_instructions2}
\end{figure*}

\begin{figure*}
    \centering
    \includegraphics[width=\textwidth]{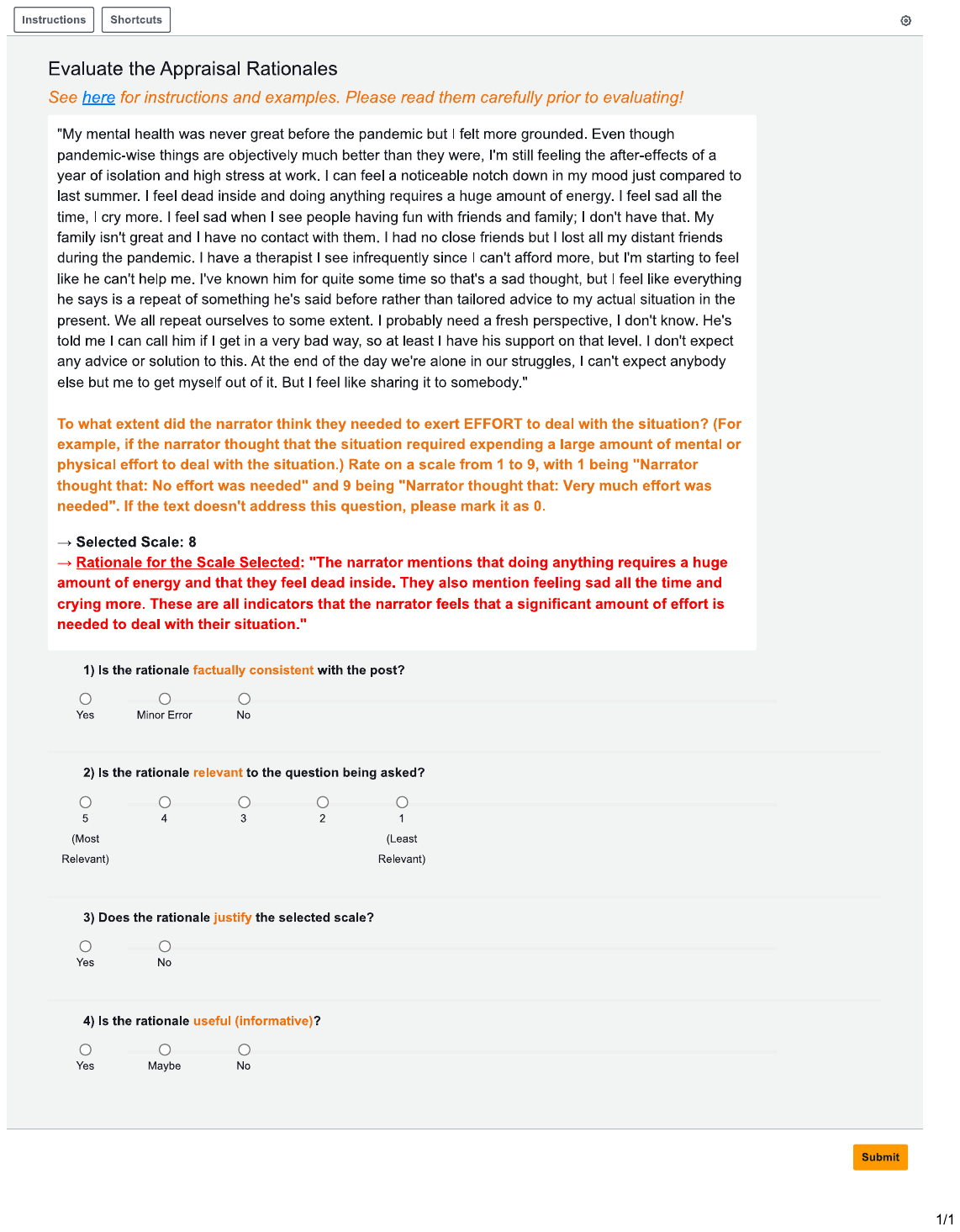}
    \caption{Task layout for the human evaluation.}
    \label{fig:human_eval_layout}
\end{figure*}

We provide the instructions given to the human evaluators of the rationales (described in \secref{subsec:human_eval}) in Figure \ref{fig:human_eval_instructions1} and Figure \ref{fig:human_eval_instructions2}. Additionally, we showcase the human evaluation task layout in Figure \ref{fig:human_eval_layout}.

\section{Why Does ChatGPT Perform (Slightly) Better Than Human Annotators in Providing Rationales?}\label{appendix:human_chatgpt_factuality}
As discussed in \secref{subsec:human_eval}, ChatGPT was scored slightly higher in terms of \textit{factuality} and \textit{usefulness} on providing natural language rationales than our human annotators, according to human evaluators. This can be attributed to ChatGPT's wordiness and extractiveness (as shown in Table \ref{tab:llm_results-rationale}), especially in cognitive emotion appraisal dimensions where the scale rating is low. As an example, we showcase in Table \ref{tab:human_chatgpt_rationale_ex} where both ChatGPT and our human annotator give the same rating for a dimension, but ChatGPT scores higher than our human experts on metrics \textit{factuality} and \textit{usefulness}.

As shown in the example, given the same Reddit post as well as the instruction to evaluate the cognitive emotion appraisal dimension \textit{orcl} (other-controllable), both our human annotator and ChatGPT give a Likert rating of $1$ indicating a small extent to which the narrator believes other people were controlling what was happening in the situation. Similarly, in their natural language rationales, both our annotator and ChatGPT express that there is no mention of other people controlling the situation in the Reddit post. However, in the post, although to a small degree, the narrator is hinting that other people are in control of the situation since their school is deciding on the mask policy for the school year. This lowers the human annotation to a mean \textit{factuality} rating of $0.25$ and a mean \textit{usefulness} rating of $0.50$. On the other hand, ChatGPT iterates and summarizes the information provided in the post, mentioning that the narrator is \textit{``seeking advice on where to find masks that are more suitable for their needs''}. The provision of such information thus leads to a higher score than our annotator, resulting in a mean rating of $0.75$ for both \textit{factuality} and \textit{usefulness}.

\clearpage
\newpage


\end{document}